\documentclass{article} 
\usepackage{iclr2027_conference,times}

\usepackage{amsmath,amsfonts,bm}

\def\eqref#1{equation~\ref{#1}}

\def\1{\bm{1}}

\DeclareMathAlphabet{\mathsfit}{\encodingdefault}{\sfdefault}{m}{sl}
\SetMathAlphabet{\mathsfit}{bold}{\encodingdefault}{\sfdefault}{bx}{n}

\usepackage{hyperref}
\usepackage{url}

\usepackage{multirow}
\usepackage{booktabs}
\usepackage{makecell}
\usepackage{graphicx}
\usepackage{caption}
\usepackage{subcaption}
\usepackage{float}
\usepackage{enumitem}
\usepackage{xcolor}
\usepackage{listings}
\lstdefinestyle{judgeprompt}{
basicstyle=\ttfamily\footnotesize,
breaklines=true,
columns=fullflexible,
keepspaces=true,
showstringspaces=false,
frame=single,
framerule=0.3pt,
xleftmargin=0.5em,
xrightmargin=0.5em
}

\title{From Judgment Quality to Downstream Utility: Rethinking LLM-as-a-Judge for Open-Ended Tasks}

\author{
\textbf{Zheng Zhang\textsuperscript{1,2}\thanks{Equal contribution.}\hspace{0.4em},
Lufei Li\textsuperscript{1}\footnotemark[1]\hspace{0.4em},
Xinyue Tan\textsuperscript{1},
Yuanhao Zeng\textsuperscript{1,2}},
Ziwei Shan\textsuperscript{1,2} \\
\textbf{    
Yexin Li\textsuperscript{2}\footnotemark[2]\hspace{0.4em},
Kan Ren\textsuperscript{1}\thanks{Corresponding authors.}} \\
\textsuperscript{1}School of Information Science and Technology, ShanghaiTech University \\
\textsuperscript{2}State Key Laboratory of General Artificial Intelligence, BIGAI \\
\texttt{\{zhangzheng2024,lilf2024,renkan\}@shanghaitech.edu.cn}
}

\iclrfinalcopy 
\begin{document}

\maketitle
\fancyhead[L]{}

\begin{abstract}
LLM-as-a-Judge is increasingly used to evaluate policy responses on open-ended tasks that lack ground-truth answers. 
Existing work often directly converts the resulting judgments into reward signals for policy training, paying limited attention to intrinsic judgment quality and largely restricting the use of Judges to training-time supervision. 
We systematically investigate judgment quality and downstream utility by examining both how judgments are elicited and how they are used.
For judgment elicitation, we vary the Judge protocol along three dimensions: verdict granularity, critique usage, and evaluation batching.
For judgment usage, beyond policy training, we extend Judge to test-time inference through Best-of-$N$ selection, Judge-guided revision, and beam search.
We find that, 
(i) Surprisingly, judgment quality and downstream utility do not always align.
(ii) Judge protocol design substantially affects both intrinsic judgment quality and downstream utility.
(iii) Judge guidance effectively converts test-time compute into performance gains, with benefits varying across inference strategies.
Our results call for a multifaceted evaluation of LLM Judges on open-ended tasks, encompassing intrinsic judgment quality, and downstream utility.
\end{abstract}

\section{Introduction}
LLM-as-a-Judge has become a widely adopted evaluation paradigm for open-ended question-answering tasks, which typically lack verifiable ground-truth answers.
Under this paradigm, a Judge assesses policy\footnote{We use \textit{policy} to represent the underlying LLM generator for a given task query.} responses against predefined rubric criteria and produces judgments indicating how well each criterion is satisfied.
These judgments provide an evaluation signal that can be further converted into numerical rewards for policy optimization.
Recent studies \citep{huang2026bootstrapping, zhang2026grad2reward, wu2026rlac, wang2026infimedorbit} increasingly use rubric-based judgments as supervision for reinforcement learning (RL) optimization.
\citet{gunjal2026rubrics} trains policies with GRPO on open-ended medical and scientific tasks using rubric-based rewards derived from a Judge’s verdicts.
\citet{zhang2026grad2reward} convert the Judge's verdict into fine-grained rewards for policy optimization.

Despite the widespread adoption of rubric-based LLM judges, two key limitations remain.
(i) \textit{Limited downstream utility.}
On open-ended tasks, Judges are primarily used to produce judgments that are converted into rewards for policy training \citep{zhou2025breaking, gunjal2026rubrics}.
The utility of judgments is mediated through policy optimization and realized only after model parameters are updated.
Policy training, however, represents only one form of downstream utility; whether Judges can also improve policy responses by directly guiding test-time inference remains underexplored.
(ii) \textit{Limited attention to intrinsic judgment quality.}
Understanding judgment quality is essential for determining whether a Judge provides reliable evaluation signals.
However, existing studies \citep{wei2026qurl, wang2026infimedorbit} focus primarily on downstream policy performance, leaving intrinsic judgment quality largely underexplored.
Moreover, whether policy performance reliably reflects the underlying judgment quality has not been systematically examined.

\begin{figure}[t]
    \centering
    \includegraphics[width=1.0\textwidth]{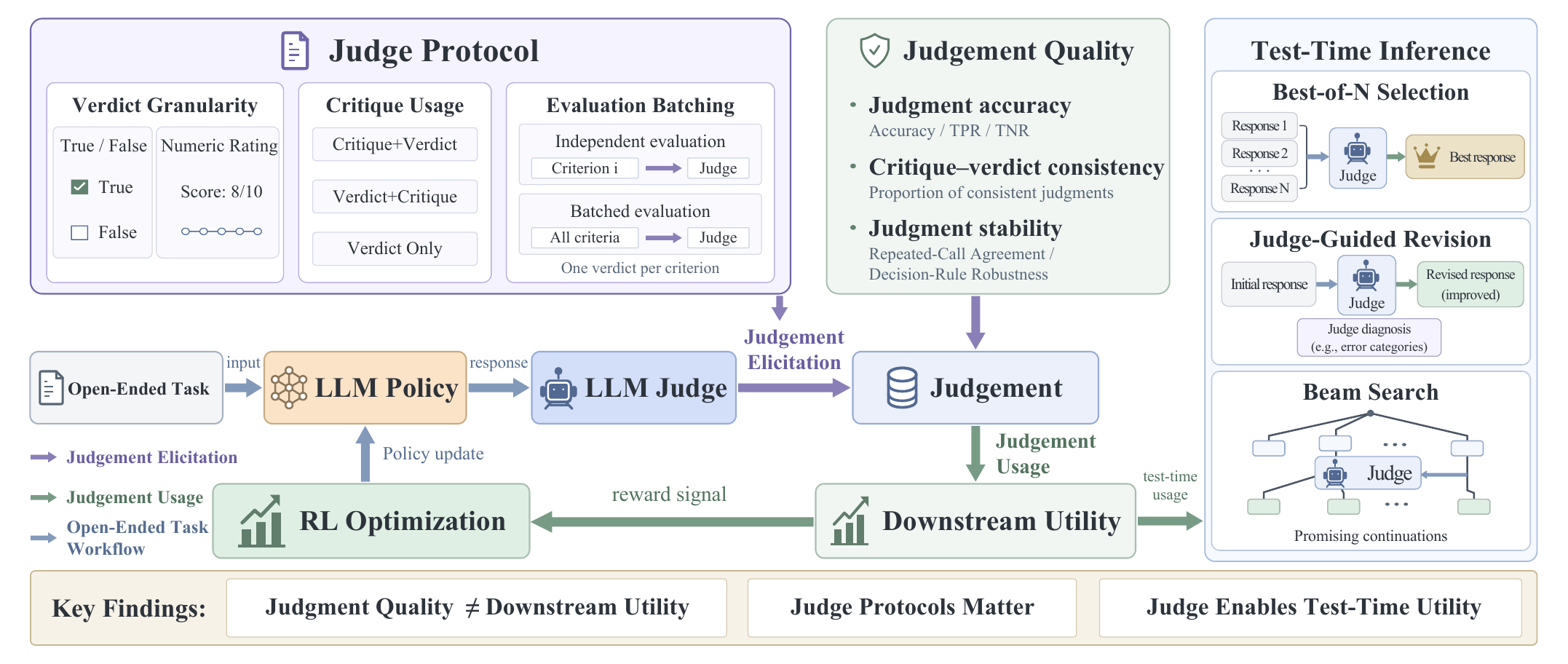}
    \vspace{-1.5em} 
    \caption{
    Overview of our framework for analyzing judgment quality and downstream utility in open-ended tasks by varying Judge protocols.
    }
    \label{fig:overview}
\end{figure}

To address these limitations, we systematically investigate judgment quality and downstream utility along two axes: how judgments are elicited and how they are used.
For judgment elicitation, we vary the Judge protocol along three dimensions: verdict granularity, critique usage, and evaluation batching.
We assess each protocol's intrinsic judgment quality through judgment accuracy, critique--verdict consistency, and stability.
For judgment usage, we examine the Judge's utility beyond training by applying it to test-time inference through Best-of-$N$ selection, Judge-guided revision, and beam search.
We conduct this investigation on open-ended question-answering tasks spanning the medical and scientific domains, using multiple policy and Judge models.
The overall pipeline is illustrated in Figure~\ref{fig:overview}.

Our investigation yields three main findings:
\begin{itemize}[leftmargin=1.0em, labelsep=0.3em, nosep]
\item \textit{Judgment quality and downstream utility do not always align.}
Higher-quality judgments do not necessarily lead to better policies after training.
\item \textit{Judge protocols matter.}
Judge protocols across verdict granularity, critique usage, and evaluation batching substantially influence judgment quality and policy optimization outcomes.
\item \textit{Test-time Judge guidance provides downstream utility.} Best-of-$N$ selection, Judge-guided revision, and beam search improve policy responses, with gains varying across strategies.

\end{itemize}

\section{Related Work}
\paragraph{Judge for Open-Ended Tasks.}
LLM Judges are increasingly used to support policy optimization on open-ended question-answering tasks, such as medical consultation \citep{arora2025healthbench} and scientific question answering \citep{yifei2025researchqa}, by assessing policy responses against predefined rubric criteria and converting the resulting judgments into rewards \citep{zhou2025breaking, shao2025dr, xu2026alternating, huang2025reinforcement, jiang2026from}.
For example, \citet{wei2026qurl} train policies for open-ended question answering using Judge assessments of factual soundness and writing quality, whereas \citet{wang2026infimedorbit} optimize policies for open-ended medical dialogue using rubric-based Judge rewards.
However, these studies pay limited attention to intrinsic judgment quality and to the broader downstream utility of LLM Judges beyond policy training.

\paragraph{Analysis of Judge Usage.}
Existing analyses of LLM judge usage can be categorized by evaluation format into pairwise and pointwise settings.
Pairwise studies \citep{wei2024systematic, feng2025we, liu2025inference, qian2026who, li2026preference} employ the Judge as a preference evaluator and examine how protocol affect preference alignment.
\citet{liu2026examining} examines the effects of reasoning and non-reasoning judges on LLM alignment.
Pointwise studies \citep{yamauchi-etal-2026-empirical, siro-etal-2026-learning, shen2026rethinking} focus on settings in which a Judge evaluates each open-ended response independently against a predefined rubric.
For instance,
\citet{song2026beyond} investigate how rubric design shapes inter-Judge agreement, while \citet{siro-etal-2026-learning} analyze how rubric provenance affects Judge assessments.
However, existing analysis either focus on LLM alignment or examine the effects of rubric design, while the quality of judgments in open-ended tasks remains underexplored.

\section{Preliminary}
\subsection{Judges as Training Signal Providers for Open-Ended Tasks}

For open-ended tasks, given a query \(x\) and a policy-generated response \(o\), existing methods
\citep{bi2025reward, he-etal-2026-advancedif, shen2026rethinking}
evaluate the response using an \textit{LLM judge} together with a predefined,
query-specific rubric
$\mathcal{R}(x)=\{(c_k,w_k)\}_{k=1}^{K}$,
where $c_k$ specifies an evaluation criterion and $w_k$ denotes its
corresponding importance weight
\citep{sheng2026reinforcing, li-etal-2026-rubrichub, wang2026co}.
For each criterion $c_k$, the judge receives a structured prompt containing
the query $x$, the response $o$, and the criterion $c_k$, and produces a
binary verdict
\begin{equation}
z_k \sim p_{\mathrm{judge}}(\cdot \mid x,o,c_k),
\qquad
z_k \in \{\texttt{True},\texttt{False}\}
\end{equation}
indicating whether the response satisfies that criterion.

The criterion-level verdicts are subsequently aggregated into a normalized
reward:
\begin{equation}
r(x,o)
=
\frac{
\sum_{k=1}^{K} w_k
\mathbb{I}\!\left[z_k=\texttt{True}\right]
}{
\sum_{k=1}^{K} \max(w_k, 0)
}
\label{eq:sequence_reward}
\end{equation}
where $\mathbb{I}[\cdot]$ is the indicator function. The resulting reward $r(x,o)$ summarizes the overall quality of the policy response and can be directly used by RL algorithms, such as GRPO
\citep{shao2024deepseekmath}.

\subsection{Judge Protocols}
\label{sec:judge_protocol}
We systematically survey how prior work elicits judgments from LLM Judges and organize common protocols along three dimensions: verdict granularity, critique usage, and evaluation batching.

\paragraph{Verdict Granularity:}
Verdict granularity refers to how precisely a judge expresses the degree to which a response satisfies a given criterion. We consider two common formats:
(1) \textbf{T/F} produces a binary verdict of \texttt{True} or \texttt{False}, indicating whether the criterion is satisfied.  
(2) \textbf{Rating} adopts the $0$--$10$ scoring scheme used in prior work \citep{whitehouse2026j, zhang-etal-2025-longreward} to measure the extent to which the criterion is satisfied.

\paragraph{Critique Usage:}
Critique usage concerns whether the Judge generates a natural-language critique and, if so, whether the critique precedes or follows the verdict. This gives rise to three protocols:
(1) \textbf{Critique+Verdict} generates a natural-language critique before the verdict.
(2) \textbf{Verdict+Critique} produces a verdict followed by a critique.
(3) \textbf{Verdict Only} outputs the verdict without a critique.

\paragraph{Evaluation Batching:}
Evaluation batching captures how many rubric criteria are assessed within each Judge call. Two protocols are considered:
(1) \textbf{Independent evaluation} calls the Judge separately for each criterion to assess whether the response satisfies it.
(2) \textbf{Batched evaluation} calls the Judge once to assess the response against all criteria, producing a verdict for each.

In open-ended tasks, a common configuration evaluates each rubric criterion independently, with the Judge generating a critique followed by a binary verdict.
This configuration corresponds to \textbf{T/F}, \textbf{Critique+Verdict}, and \textbf{Independent Evaluation}.
Taking it as our reference setting, we systematically investigate the effects of varying each protocol dimension described above.
Prompt templates and implementation details for each protocol are provided in the Appendix~\ref{appendix:judge_templates}.

\section{Analytical Framework}
\label{sec:analytical_framework}
We vary Judge protocols to investigate two aspects of LLM Judges: their utility in open-ended RL training and the intrinsic quality of their judgments.
We assess RL training utility through reward signal statistics and downstream policy performance, and judgment quality through accuracy, critique--verdict consistency, and stability.

\subsection{Downstream Utility for RL Training.}
We assess RL training effectiveness from two perspectives:

\textbf{(1) Reward Signal Statistics.}
GRPO samples multiple responses to each query, forming a \textit{response group}, and uses their relative rewards to estimate advantages for policy optimization. We characterize within-group reward resolution using the following two metrics. 
Detailed calculations and examples are provided in Appendix~\ref{appendix:reward_signal_metric}.

\begin{itemize}[leftmargin=1.5em, labelsep=0.3em, nosep]
\item \textbf{Response-Level Reward Tie Rate}: the average proportion of response pairs with identical rewards within a group. A \textit{lower} value indicates the Judge \textit{distinguishes more response pairs}.
\item \textbf{Group-Level Reward Uniformity Rate}: the proportion of response groups in which all responses receive the same reward.  
A \textit{lower} value indicates \textit{more groups provide learning signals}. 
\end{itemize}
For example, for a response group with rewards $[0.8,0.8,0.8,0.8,0.8,0.8,0.6,0]$ yield a relatively high Response-Level Reward Tie Rate of $15/28=53.6\%$. However, the group is not entirely uniform and therefore does not count toward the Group-Level Reward Uniformity Rate.

\textbf{(2) Downstream Task Performance.}
We evaluate downstream utility in two open-ended domains: medicine and science.
For medicine, we train on RaR-Medicine and evaluate on HealthBench and the held-out RaR-Medicine test set; for science, we train on RaR-Science and evaluate on ResearchQA and the held-out RaR-Science test set.
Each query is associated with multiple query-specific rubric criteria for assessing policy responses.
We optimize Qwen2.5-1.5B-Instruct and Qwen3-1.7B with GRPO, sampling eight responses per query and computing Judge-derived rewards according to Eq.~\eqref{eq:sequence_reward} using either Qwen2.5-3B-Instruct or GPT-OSS-20B as the training-time Judge. Additional results with policy Qwen2.5-7B-Instruct are provided in Appendix~\ref{appendix:larger_policies}.
We then evaluate the trained policies using GPT-OSS-120B, a stronger external grader used solely for evaluation, and report the average rubric-based score on each test set.

\subsection{Judgment Quality.}
To examine judgment quality directly, we collect policy responses and the corresponding judgments during training to construct a static dataset.
Given fixed responses, we analyze the judgments along three dimensions: judgment accuracy, critique--verdict consistency, and judgment stability.

\textbf{(1) Judgment Accuracy.}
To assess how accurately the Judge determines whether a response satisfies a given criterion, we first construct gold labels using three stronger models: GPT-OSS-120B, GPT-4.1, and DeepSeek-V3.2. 
We use majority vote to obtain gold labels.
We then compare each Judge's verdicts with the corresponding gold labels and report the following metrics.

\begin{itemize}[leftmargin=1.5em, labelsep=0.3em, nosep]
\item \textbf{Accuracy}: the proportion of the Judge's verdicts that match the gold labels. 
\textit{A higher value indicates stronger overall agreement.}
\item \textbf{TPR (True Positive Rate)}: the proportion of cases with positive gold labels that the Judge classifies as positive. 
\textit{A higher value indicates the Judge identifies more positive cases.}
\item \textbf{TNR (True Negative Rate)}: the proportion of cases with negative gold labels that the Judge classifies as negative. 
\textit{A higher value indicates the Judge identifies more negative cases.}
\end{itemize}

\textbf{(2) Critique--Verdict Consistency.}
To assess whether the Judge's critique logically supports its verdict, we use GPT-OSS-120B as a separate auditor. Given the original response, rubric criterion, and the Judge's critique and verdict, the auditor determines whether the critique supports the verdict (auditor prompt template in Appendix~\ref{appendix:consistency_auditor_prompt}). We report the proportion of judgments deemed consistent. 
\textit{A higher value indicates stronger consistency between critiques and verdicts.}

\textbf{(3) Judgment Stability.}
To examine the repeatability of the Judge's judgments and their robustness to decision-rule changes, we use two metrics:  \begin{itemize}[leftmargin=1.5em, labelsep=0.3em, nosep]
\item \textbf{Repeated-Call Agreement}: 
we judge each fixed response--criterion pair five times under same configuration. 
To measure verdict repeatability without privileging any single call, we average verdict agreement over ten pairwise comparisons.
\textit{A higher value indicates greater repeatability.}
\item \textbf{Decision-Rule Robustness}:
For each fixed response--criterion pair, we obtain verdicts under conservative, neutral, and permissive decision rules (see Appendix~\ref{appendix:decision_rule_prompts} for Judge templates). 
We report the proportion of response--criterion instances for which the verdict remains unchanged across all three rules.
\textit{A higher value indicates greater robustness to decision-rule changes.}
\end{itemize}

The metrics above are defined for binary T/F verdicts and can be extended to Rating verdicts. 
Detailed formulas, illustrative examples, and the implementations are provided in Appendix~\ref{appendix:metric_definitions}.

\section{Verdict Granularity}
In this section, we examine how verdict granularity affects intrinsic judgment quality and downstream utility.
We compare the two verdict formats introduced in Section~\ref{sec:judge_protocol}: \textit{T/F} outputs a True/False verdict, whereas \textit{Rating} outputs a score from 0 to 10.
All other protocol components and training configurations are held fixed.

\subsection{Downstream Utility in RL Training}
\paragraph{Rating improves reward resolution and generally yields stronger downstream performance, especially with a more capable Judge.}
As shown in Table~\ref{tab:policy_judge_verdict_form_rl}, Rating achieves a higher mean test score than T/F in 10 of the 16 matched comparisons.
Its advantage is most consistent with GPT-OSS-20B as Judge, where it performs better in seven of eight comparisons.
The reward statistics in Figure~\ref{fig:reward_resolution} and ~\ref{fig:reward_resolution_rating} help explain this overall advantage: Rating reduces the Response-Level Reward Tie Rate in all eight settings and the Group-Level Reward Uniformity Rate in seven.
By assigning fine-grained scores, Rating distinguishes responses that T/F treats as equivalent and enables more response groups to provide non-zero relative advantages for GRPO.
Together, these results suggest that Rating provides more informative optimization signals.

\begin{table}[H]
\centering
\caption{Test performance of different policy models optimized with different training-time judges and verdict forms. Values are mean $\pm$ standard deviation over repeated runs. Bold indicates the higher mean within each matched setting.}
\label{tab:policy_judge_verdict_form_rl}
\resizebox{\textwidth}{!}{
\begin{tabular}{lllcccc}
\toprule
\textbf{Policy}
& \textbf{\makecell{Training-Time\\Judge}}
& \textbf{\makecell{Verdict\\Form}}
& \textbf{\makecell{Health\\Bench}}
& \textbf{\makecell{RaR-\\Medicine}}
& \textbf{\makecell{Research\\QA}}
& \textbf{\makecell{RaR-\\Science}} \\
\midrule

\multirow{4}{*}{\makecell{Qwen2.5-1.5B-Instruct}}& \multirow{2}{*}{\makecell{Qwen2.5-3B-Instruct}}
& T/F    &  \textbf{0.140 $\pm$ 0.006}&  \textbf{0.266 $\pm$ 0.004}&  $0.371 \pm 0.021$&  $0.302 \pm 0.003$\\
&
& Rating &  $0.139 \pm 0.007$&  $0.255 \pm 0.014$&  \textbf{0.400 $\pm$ 0.071}&  \textbf{0.333 $\pm$ 0.019}\\
\cmidrule(lr){2-7}
& \multirow{2}{*}{gpt-oss-20b}
& T/F    &  $0.158 \pm 0.014$&  $0.280 \pm 0.005$&  $0.383 \pm 0.005$&  $0.334 \pm 0.006$\\
&
& Rating &  \textbf{0.164 $\pm$ 0.003}&  \textbf{0.285 $\pm$ 0.006}&  \textbf{0.402 $\pm$ 0.028}&  \textbf{0.349 $\pm$ 0.005}\\

\midrule

\multirow{4}{*}{Qwen3-1.7B}
& \multirow{2}{*}{\makecell{Qwen2.5-3B-Instruct}}
& T/F    &  \textbf{0.260 $\pm$ 0.005}&  $0.313 \pm 0.008$&  \textbf{0.558 $\pm$ 0.006}&  \textbf{0.516 $\pm$ 0.005}\\
&
& Rating &  $0.252 \pm 0.008$&  \textbf{0.315 $\pm$ 0.004}&  $0.555 \pm 0.013$&  $0.505 \pm 0.002$\\
\cmidrule(lr){2-7}
& \multirow{2}{*}{gpt-oss-20b}
& T/F    &  $0.261 \pm 0.004$&  $0.321 \pm 0.004$&  $0.557 \pm 0.007$&  \textbf{0.522 $\pm$ 0.007}\\
&
& Rating &  \textbf{0.269 $\pm$ 0.003}&  \textbf{0.326 $\pm$ 0.005}&  \textbf{0.589 $\pm$ 0.014}&  $0.520 \pm 0.011$\\

\bottomrule
\end{tabular}
}
\end{table}

\subsection{Judgment Quality}

\begin{figure}[htbp]
  \centering
  \begin{minipage}[c]{0.48\linewidth}
    \centering
    \includegraphics[width=\linewidth]{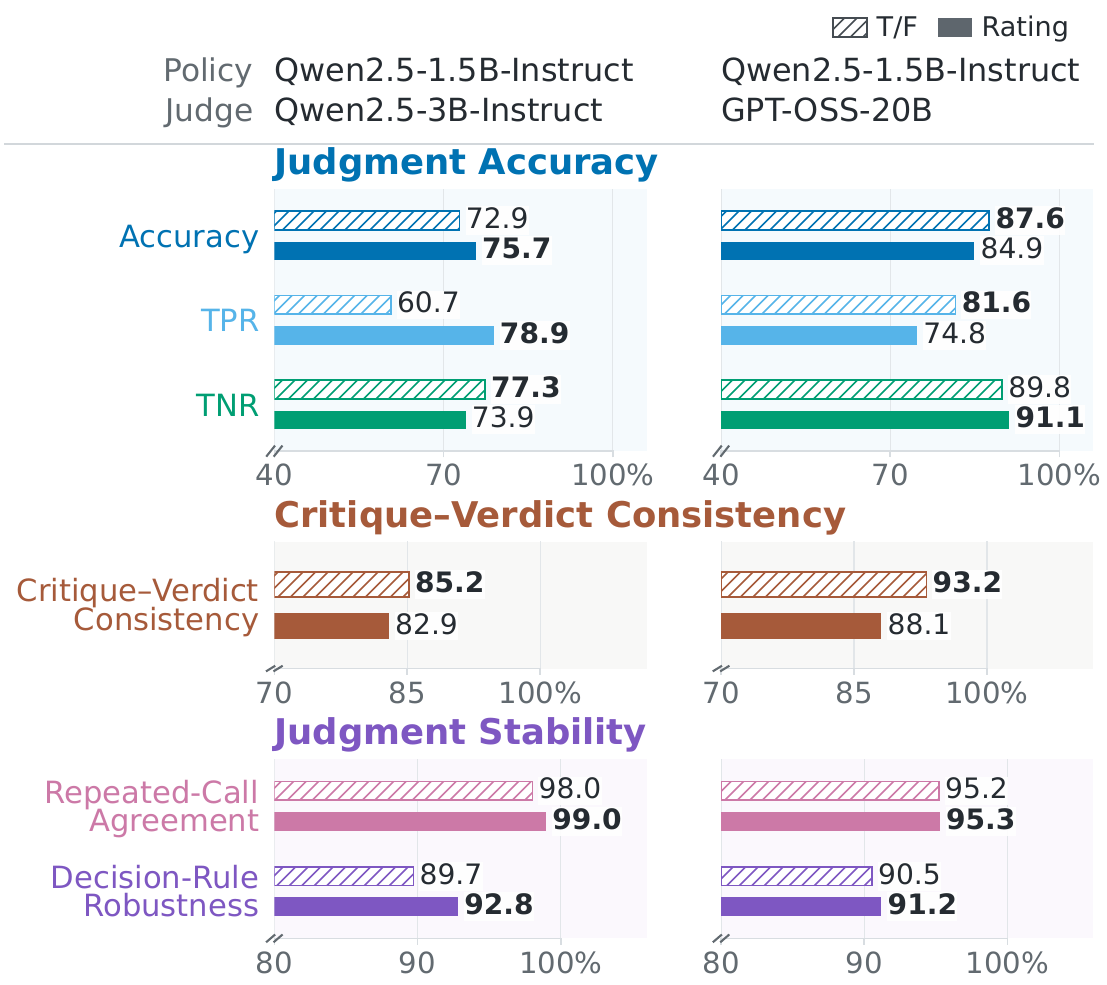}
    \vspace{-1.5em}
    \caption{Judgemet quality of different verdict forms on Medicine. }
    \label{fig:verdict_granularity}
  \end{minipage}
  \hfill
  \begin{minipage}[c]{0.48\linewidth}
\paragraph{Rating changes the judgment profile rather than uniformly improving
judgment quality.}
For Judgment Accuracy, Rating has Judge-dependent effects (Figures\ref{fig:appendix_verdict_granularity_full} and~\ref{fig:verdict_granularity}).
With Qwen2.5-3B-Instruct, it increases Accuracy in three and TPR, while decreasing TNR.
With GPT-OSS-20B, it decreases Accuracy and TPR in all four settings and TNR in three.
For Critique--Verdict Consistency, T/F outperforms Rating in seven of eight settings.
For Judgment Stability, Rating achieves higher Repeated-Call Agreement and Decision-Rule Robustness in seven of eight settings. Overall, its stronger downstream performance aligns with finer reward resolution and greater stability, but not uniformly higher Judgment Accuracy or Critique--Verdict Consistency.
  \end{minipage}
\end{figure}

\begin{quote}
\textbf{Summary.}
\textit{
Rating provides finer-grained reward signals and supports more effective policy training than T/F judgments, despite showing no consistent advantage in intrinsic judgment quality.
}
\end{quote}

\section{Critique Usage}
\label{sec:critique_usage}

In this section, We examine how critique usage affects Judge decisions and downstream policy optimization.
We compare Critique+Verdict, Verdict Only, and Verdict+Critique, as defined in Section~\ref{sec:judge_protocol}.
All other protocol components and training configurations are held fixed.

\subsection{Downstream Utility in RL Training}
\paragraph{Judges using a Verdict+Critique output order generally yield the strongest downstream policy performance.}
Table~\ref{tab:critique_output_rl} shows that Verdict+Critique achieves the highest mean performance in 12 of the 16 settings, compared with only three settings each for Critique+Verdict and Verdict Only. 
Its advantage is strongest with Qwen2.5-3B-Instruct. One possible explanation is that an inaccurate critique can anchor the subsequent verdict for a less capable Judge, whereas verdict-first generation prevents the critique from influencing the decision.
Meanwhile, the reward-resolution statistics in Appendix~\ref{appendix:critique_reward_resolution} show that no critique protocol consistently improves reward resolution.
Overall, the results favor Verdict+Critique for policy optimization, particularly with less capable Judges.

\begin{table}[H]
\centering
\caption{Test performance of RL optimization with different critique and verdict-position settings across training-time judges. 
Best mean per column is bold; second-best is underlined.
}
\label{tab:critique_output_rl}
\resizebox{\textwidth}{!}{
\begin{tabular}{lllcccc}
\toprule
\textbf{Policy}
& \textbf{Training-Time Judge}
& \textbf{\makecell{Judge\\Output}}
& \textbf{\makecell{Health\\Bench}}
& \textbf{\makecell{RaR-\\Medicine}}
& \textbf{\makecell{Research\\QA}}
& \textbf{\makecell{RaR-\\Science}} \\
\midrule

\multirow{3}{*}{Qwen2.5-1.5B-Instruct}
& \multirow{3}{*}{Qwen2.5-3B-Instruct}
& Critique+Verdict    &  $0.140 \pm 0.006$&  \underline{0.266 $\pm$ 0.004}&  $0.371 \pm 0.021$&  $0.302 \pm 0.003$\\
&
& Only Verdict &  \underline{0.149 $\pm$ 0.007}&  $0.257 \pm 0.012$&  \underline{0.398 $\pm$ 0.019}&  \underline{0.340 $\pm$ 0.004}\\
&
& Verdict+Critique & \textbf{0.153 $\pm$ 0.005}&  \textbf{0.275 $\pm$ 0.009}&  \textbf{0.414 $\pm$ 0.018}&  \textbf{0.346 $\pm$ 0.002}\\
\cmidrule(lr){2-7}
& \multirow{3}{*}{gpt-oss-20b}
& Critique+Verdict    &  \underline{0.158 $\pm$ 0.014}&  $0.280 \pm 0.005$&  \textbf{0.383 $\pm$ 0.005}&  \textbf{0.334 $\pm$ 0.006}\\
&
& Only Verdict &  $0.156 \pm 0.004$&  \textbf{0.285 $\pm$ 0.002}&  \underline{0.367 $\pm$ 0.006}&  \underline{0.332 $\pm$ 0.004}\\
&
& Verdict+Critique &  \textbf{0.160 $\pm$ 0.010}&  \underline{0.282 $\pm$ 0.008}&  $0.366 \pm 0.002$&  \textbf{0.334 $\pm$ 0.002}\\
\midrule

\multirow{3}{*}{Qwen3-1.7B}
& \multirow{3}{*}{Qwen2.5-3B-Instruct}
& Critique+Verdict    &  \underline{0.260 $\pm$ 0.005}&  \underline{0.313 $\pm$ 0.008}&  \underline{0.558 $\pm$ 0.006}&  \underline{0.516 $\pm$ 0.005}\\
&
& Only Verdict &  $0.254 \pm 0.003$&  \textbf{0.324 $\pm$ 0.014}&  $0.553 \pm 0.005$&  \underline{0.516 $\pm$ 0.003}\\
&
& Verdict+Critique &  \textbf{0.261 $\pm$ 0.010}&  \textbf{0.324 $\pm$ 0.003}&  \textbf{0.572 $\pm$ 0.006}&  \textbf{0.528 $\pm$ 0.003}\\
\cmidrule(lr){2-7}
& \multirow{3}{*}{gpt-oss-20b}
& Critique+Verdict    &  \underline{0.261 $\pm$ 0.004}&  $0.321 \pm 0.004$&  $0.557 \pm 0.007$&  \textbf{0.522 $\pm$ 0.007}\\
&
& Only Verdict &  \textbf{0.262 $\pm$ 0.004}&  \underline{0.323 $\pm$ 0.010}&  \underline{0.561 $\pm$ 0.007}&  $0.515 \pm 0.002$\\
&
& Verdict+Critique &  \underline{0.261 $\pm$ 0.005}&  \textbf{0.325 $\pm$ 0.002}&  \textbf{0.571 $\pm$ 0.014}&  \underline{0.521 $\pm$ 0.009}\\

\bottomrule
\end{tabular}
}
\end{table}

\subsection{Judgment Quality}

\begin{figure}[htbp]
  \centering
  \begin{minipage}[c]{0.48\linewidth}
    \centering
    \includegraphics[width=\linewidth]{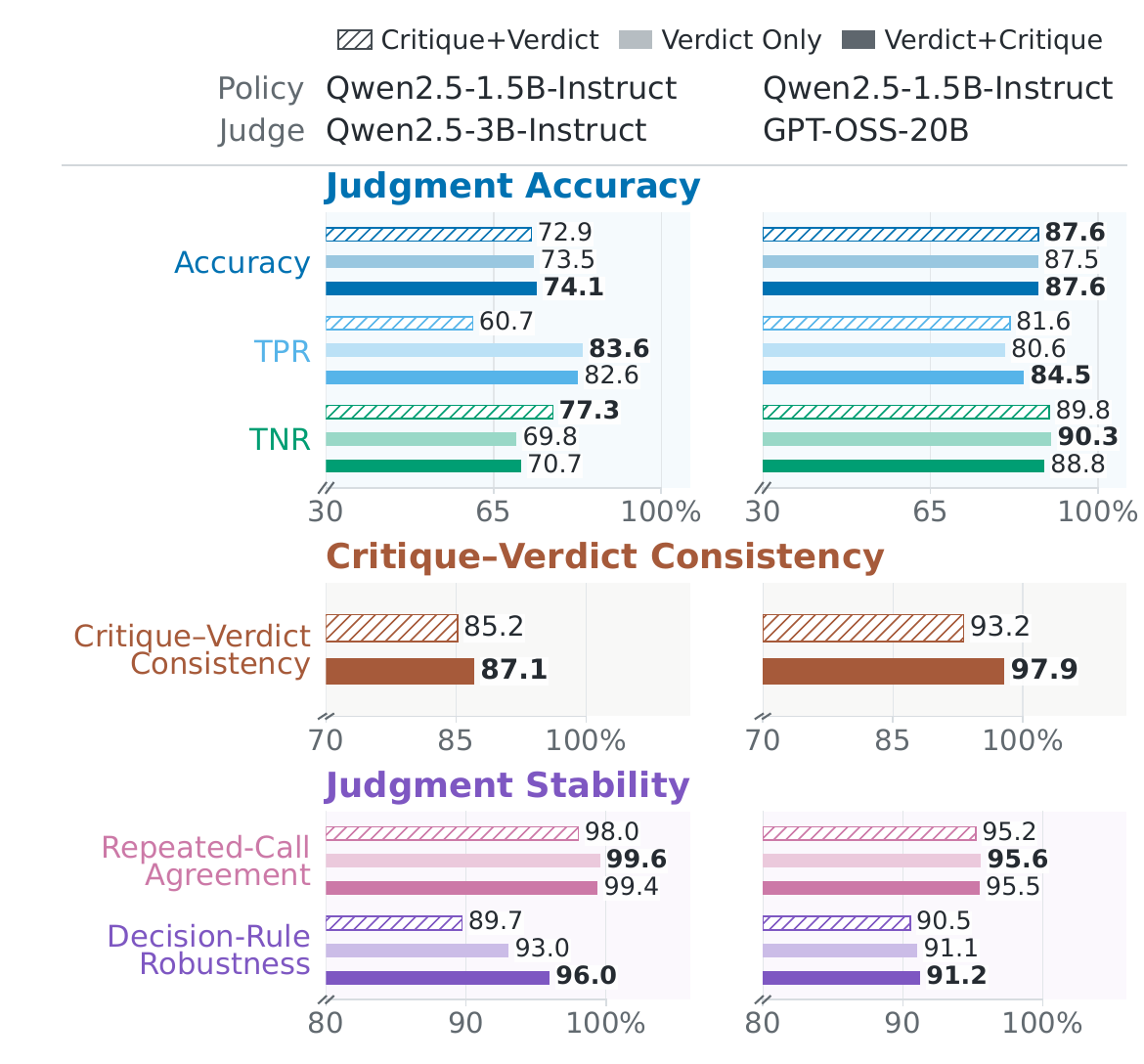}
    \vspace{-1.5em}
    \caption{Judgemet quality of different critique and verdict settings on Medicine.}
    \label{fig:critique_usage_qwen25}
  \end{minipage}
  \hfill
  \begin{minipage}[c]{0.48\linewidth}
    \centering
    \includegraphics[width=\linewidth]{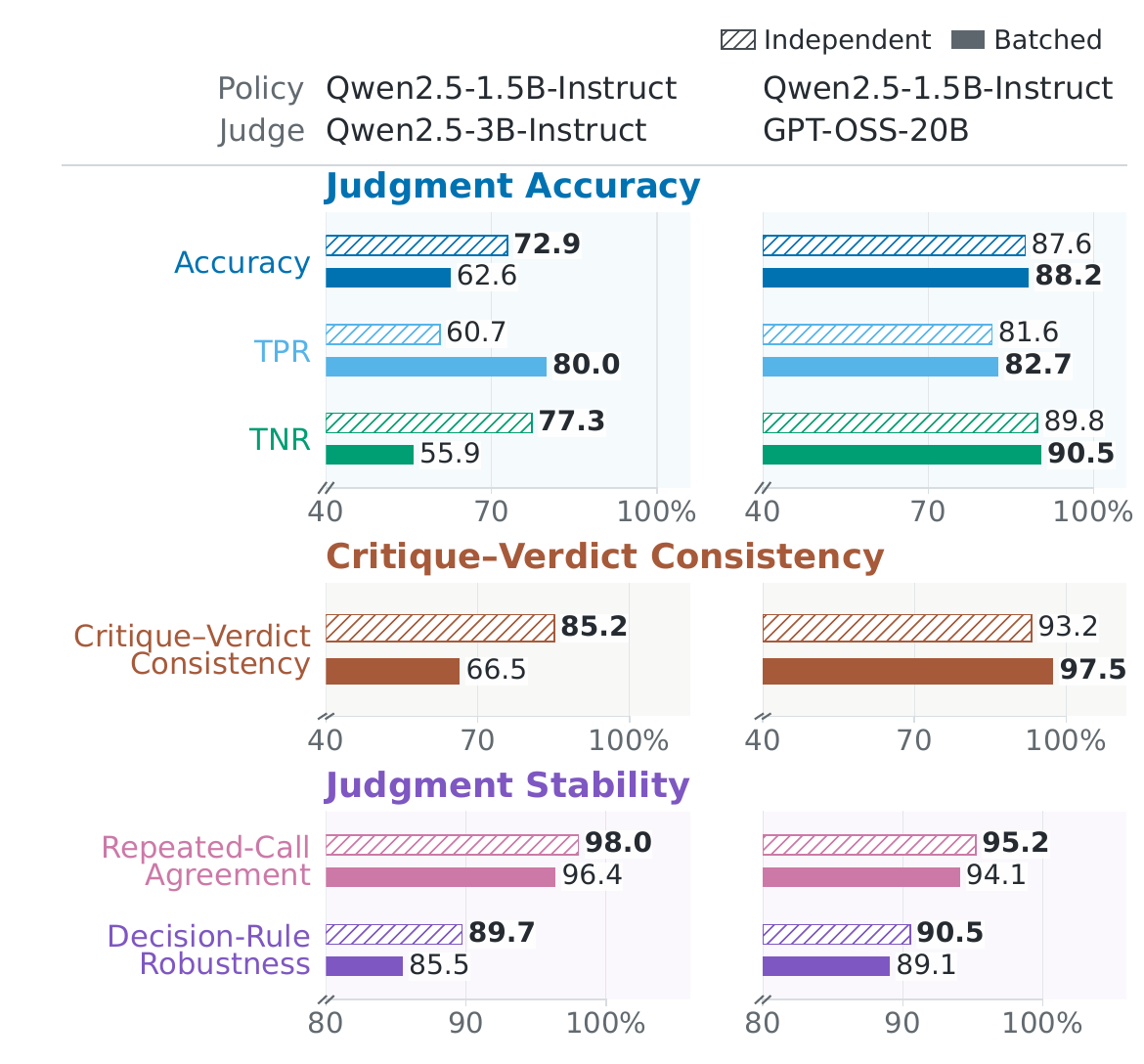}
    \vspace{-1.5em}
    \caption{Judgemet quality of single and batching settings on Medicine.}
    \label{fig:evaluation_batching_qwen25}
  \end{minipage}
\end{figure}

\paragraph{Verdict+Critique provides the strongest overall judgment quality.}
For Judgment Accuracy, Verdict+Critique achieves the highest Accuracy in five of eight settings (Figure~\ref{fig:appendix_critique_usage_full} and ~\ref{fig:critique_usage_qwen25}).  Relative to ritique+Verdict, it increases TPR but decreases TNR, indicating a shift toward more positive judgments rather than uniform improvement across error types. 
For Critique--Verdict Consistency, Verdict+Critique outperforms
Critique+Verdict in seven of eight settings, indicating that placing the critique after the verdict generally improves their logical alignment. 
For Judgment Stability, Verdict Only achieves the highest repeatability, while Verdict+Critique  achieves the highest Decision-Rule Robustness in seven of eight settings. 
Overall, Verdict+Critique provides a better balance of Accuracy, Consistency, and Stability. Together with its stronger downstream policy performance, this shows that critique--verdict ordering affects both intrinsic judgment quality and downstream utility.

\begin{quote}
\textbf{Summary.}
\textit{
For critique usage, Verdict+Critique provides the best overall balance across judgment quality dimensions, including accuracy, consistency, and stability, while also yielding the strongest downstream policy performance.
}
\end{quote}

\section{Evaluation Batching}

In this section, we examine how evaluation batching affects judgment quality and downstream utility in RL training.
We compare Independent evaluation, which evaluates each rubric criterion in a separate Judge call, with Batched evaluation, which evaluates all criteria in a single call.

\subsection{Downstream Utility in RL Training}
\label{sec:evaluation_batching}

\paragraph{Batched evaluation improves efficiency at the cost of reward resolution and downstream performance.}
Table~\ref{tab:single_group_rl} shows that Batched Evaluation achieves a higher mean performance in only six of the 16 matched comparisons. The reward statistics in Appendix~\ref{appendix:batching_reward_resolution} show that batching increases the Response-Level Reward Tie Rate in seven of eight settings and the Group-Level Reward Uniformity Rate in all eight, reducing response-level discrimination and leaving fewer groups with non-zero relative advantages for
GRPO.

\vspace{-0.4em}
\begin{table}[H]
\centering
\caption{Test performance of Independent evaluation and Batched evaluation. Values are mean $\pm$ standard deviation over repeated runs. 
}
\vspace{-0.5em}
\label{tab:single_group_rl}
\resizebox{\textwidth}{!}{
\begin{tabular}{lllcccc}
\toprule
\textbf{Policy}
& \textbf{Training-Time Judge}
& \textbf{\makecell{Judge\\Output}}
& \textbf{\makecell{Health\\Bench}}
& \textbf{\makecell{RaR-\\Medicine}}
& \textbf{\makecell{Research\\QA}}
& \textbf{\makecell{RaR-\\Science}} \\
\midrule

\multirow{4}{*}{Qwen2.5-1.5B-Instruct}
& \multirow{2}{*}{Qwen2.5-3B-Instruct}
& Independent&  $0.140 \pm 0.006$&  \textbf{0.266 $\pm$ 0.004}&  \textbf{0.371 $\pm$ 0.021}&  $0.302 \pm 0.003$\\
&
& Batched&  \textbf{0.143 $\pm$ 0.005}&  $0.239 \pm 0.005$&  $0.332 \pm 0.006$&  \textbf{0.306 $\pm$ 0.005}\\
\cmidrule(lr){2-7}
& \multirow{2}{*}{gpt-oss-20b}
& Independent&  \textbf{0.158 $\pm$ 0.014}&  \textbf{0.280 $\pm$ 0.005}&  \textbf{0.383 $\pm$ 0.005}&  $0.334 \pm 0.006$\\
&
& Batched&  $0.155 \pm 0.013$&  $0.270 \pm 0.003$&  $0.362 \pm 0.008$&  \textbf{0.335 $\pm$ 0.007}\\

\midrule

\multirow{4}{*}{Qwen3-1.7B}
& \multirow{2}{*}{Qwen2.5-3B-Instruct}
& Independent&  \textbf{0.260 $\pm$ 0.005}&  $0.313 \pm 0.008$&  \textbf{0.558 $\pm$ 0.006}&  \textbf{0.516 $\pm$ 0.005}\\
&
& Batched&  $0.252 \pm 0.007$&  \textbf{0.319 $\pm$ 0.003}&  $0.547 \pm 0.003$&  $0.494 \pm 0.005$\\
\cmidrule(lr){2-7}
& \multirow{2}{*}{gpt-oss-20b}
& Independent&  \textbf{0.261 $\pm$ 0.004}&  $0.321 \pm 0.004$&  $0.557 \pm 0.007$&  \textbf{0.522 $\pm$ 0.007}\\
&
& Batched&  $0.256 \pm 0.005$&  \textbf{0.323 $\pm$ 0.011}&  \textbf{0.562 $\pm$ 0.006}&  $0.521 \pm 0.013$\\

\bottomrule
\end{tabular}
}
\end{table}

\vspace{-1.7em}

\begin{figure}[H]
\centering
\begin{minipage}[c]{0.48\linewidth}
\centering
\includegraphics[width=\linewidth]{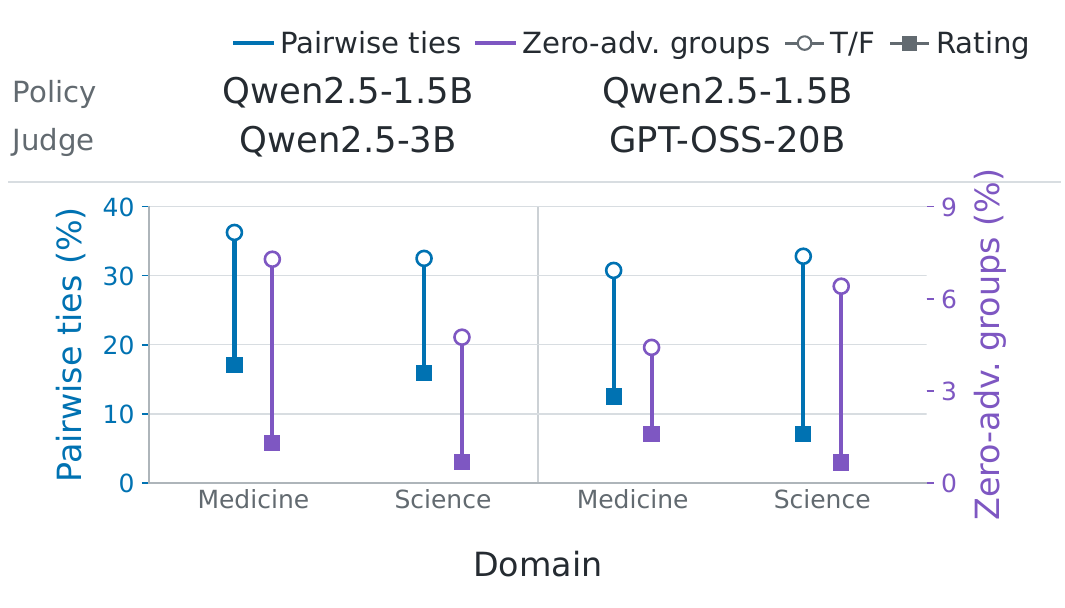}
\vspace{-1.2em}
\caption{Reward signal statistics.}
\label{fig:reward_resolution_rating}
\end{minipage}
\hfill
\begin{minipage}[c]{0.48\linewidth}
\centering
\includegraphics[width=\linewidth]{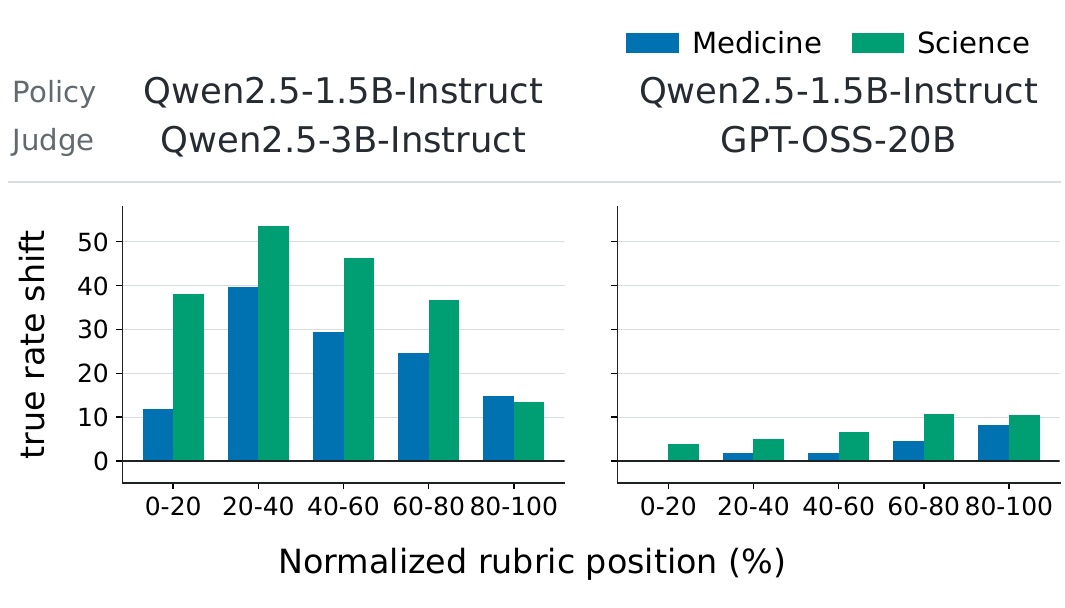}
\vspace{-1.2em}
\caption{Position bias.}
\label{fig:pos_bias}
\end{minipage}
\end{figure}

\vspace{-0.4em}

\subsection{Judgment Quality}

\paragraph{Batched evaluation generally weakens judgment quality, especially for the smaller Judge.}
For Judgment Accuracy, batched evaluation reduces accuracy in seven of eight settings (Figures \ref{fig:appendix_evaluation_batching_full} and~\ref{fig:evaluation_batching_qwen25}). With Qwen2.5-3B-Instruct, batched evaluation consistently increases TPR while decreasing TNR, indicating
a systematic shift toward more positive judgments, whereas its effects on GPT-OSS-20B are smaller. 
For Critique--Verdict Consistency, batched evaluation produces substantial declines with Qwen2.5-3B-Instruct, but mixed changes with GPT-OSS-20B, indicating that jointly processing multiple criteria can disrupt critique--verdict alignment for less capable Judges. 
For Judgment Stability, batched evaluation lowers Repeated-Call Agreement in seven of eight settings and Decision-Rule Robustness in six, suggesting greater sensitivity to sampling variation and decision rules.
Overall, batched evaluation reduces evaluation cost but generally compromises intrinsic judgment quality and downstream utility, with stronger Judges mitigating these effects.

\paragraph{Batched Evaluation makes the smaller Judge more permissive and introduces position bias.}
Figure~\ref{fig:appendix_pos_bias_full} and ~\ref{fig:pos_bias} reports the change from Independent to Batched Evaluation in True-Verdict Rate, defined as the proportion of criteria judged as True at each normalized rubric position. A positive value indicates that batching makes the Judge more likely to output True.
With Qwen2.5-3B-Instruct, the True-Verdict Rate increases at every position, with substantially larger increases in the early and middle positions than near the end of the rubric.
Thus, batching not only makes the smaller Judge more permissive overall, but also affects its decisions differently across rubric positions.
In contrast, GPT-OSS-20B exhibits much smaller changes and is therefore less sensitive to batching.

\begin{quote}
\textbf{Summary.}
\textit{Batched evaluation degrades judgment quality and causes position-dependent shifts, with limited downstream impact.}
\end{quote}

\section{downstream utility for Test-Time inference}

Beyond training rewards, we investigate whether an LLM Judge can guide test-time computation to improve policy performance without updating policy parameters. We consider three approaches: (i) \textit{Best-of-$N$ selection}, where the Judge selects the best among multiple complete policy responses; (ii) \textit{Judge-guided revision}, where the Judge provides feedback to improve an existing policy response; and (iii) \textit{Beam search}, where the Judge scores policy-generated partial responses and selects promising prefixes for continued generation.

\subsection{Best-of-N}

\textbf{Setup.} For each question, a frozen policy samples $N\in\{4,8,20,40\}$ responses. The Judges (GPT-OSS-20B or Qwen3-30B-A3B-Instruct) assign rewards and select the highest-scoring response. An independent GPT-OSS-120B test grader evaluates the selected response. We compare this selection with \textit{Direct} (one response), \textit{Average-of-$N$} (mean score across responses), and \textit{Oracle@$N$} (maximum score, representing the best possible selection). Full details are provided in Appendix~\ref{appendix:best_of_n_procedure}.

\textbf{Judge-guided Best-of-$N$ selection yields larger gains as $N$ increases.}
Figure~\ref{fig:best_and_revision} shows that both Judges select responses that outperform Direct in all 32 settings. 
As $N$ increases, Average-of-$N$ remains close to Direct, whereas Judge-selected responses improve steadily: their mean gain over Direct rises from 7.8 points at $N=4$ to 17.2 points at $N=40$. 
This suggests that the Judges can identify higher-quality responses when given more options. Figures~\ref{fig:appendix_best_of_n_full} provide the complete Best-of-$N$ results.

\subsection{Judge-Guided Revision}

\textbf{Setup.} We test whether one round of Judge feedback can help a frozen policy improve its response. For each question, the policy generates a Direct response. The Judge examines the question, response, and rubric, then selects one to three improvement codes from a fixed list of eight. In a separate rubric-free call, it uses the question, response, and selected codes to write concrete revision feedback. The policy then generates one revised response, which is evaluated by an independent GPT-OSS-120B test grader. Full details are provided in Appendix~\ref{appendix:revision_diagnosis_prompt}.

\textbf{Judge-guided revision improves policy responses.}
Figure~\ref{fig:best_and_revision} shows that, compared with the policy's Direct responses, a single Judge-guided revision improves the mean test-grader score in all eight settings, by 1.1 to 22.8 points.
Qwen3-30B-A3B yields larger gains than GPT-OSS-20B in every matched comparison, indicating that the effectiveness of test-time revision depends on the Judge providing guidance.
Figures~\ref{fig:appendix_best_of_n_full} provide the complete judge-guided revision results.

\subsection{Beam Search}

\textbf{Setup.} Beam search alternates between policy generation and Judge selection. At each step, the policy generates a total of $N$ continuations from the retained prefixes. Given the question, each partial response, and the rubric, the Judge assigns each continuation a 0--100 score reflecting its potential to become a high-quality complete response. The top $B$ unfinished prefixes are retained for the next step, while completed responses are collected for final Judge selection. We evaluate $(N,B)\in\{(4,1),(8,2),(20,5)\}$, where $N$ is the number of continuations generated per step and $B$ is the number of retained prefixes. The policy sees only the question and its own prefix, not the rubric or Judge scores. Further details and the Judge prompt are provided in Appendix~\ref{appendix:beam_search_procedure}.

\textbf{Judge-guided beam search improves with larger budgets but does not consistently outperform direct generation.}. Table~\ref{tab:beam_search} shows that larger search budgets improve performance in every setting. Beam search outperforms Direct in most comparisons but underperforms for Llama-3.1-8B on RaR-Science, particularly at smaller budgets. Thus, Judge guidance benefits from additional computation, although its gains over direct generation depend on the policy and domain.

\begin{quote}
\textbf{Summary.}
\textit{
Judge guidance effectively converts additional test-time computation into performance gains: 
Best-of-$N$ selection scales consistently with $N$, 
Judge-guided revision improves responses across all settings, and 
beam search benefits from larger budgets but remains sensitive to the policy and domain.
}
\end{quote}

\begin{table*}[t]
  \centering

  \begin{minipage}[t]{0.48\textwidth}
    \vspace{0pt}
    \centering
    \includegraphics[width=\linewidth]{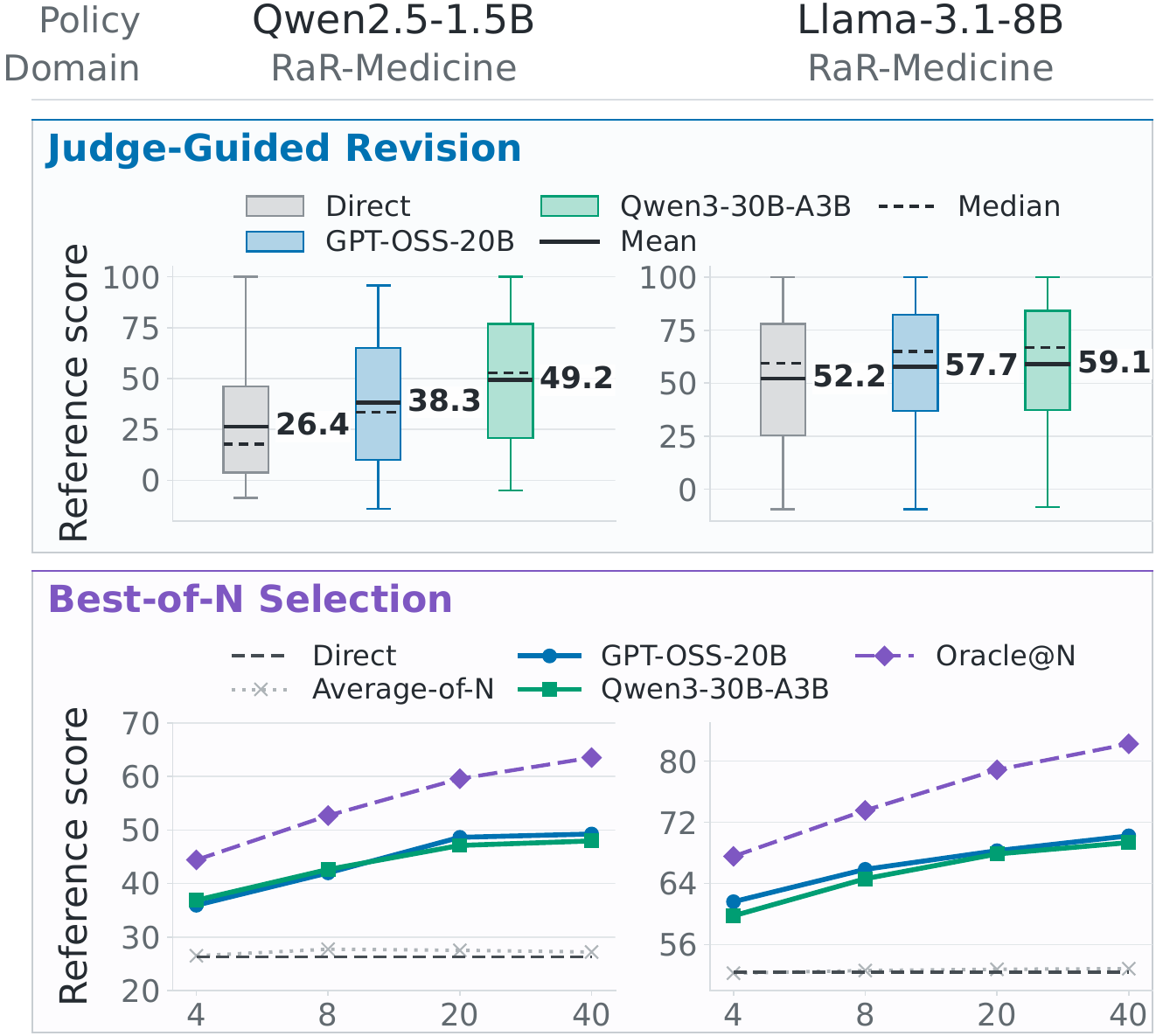}
    \vspace{-1.2em}
    \captionof{figure}{Revision performance on Medicine.}
    \label{fig:best_and_revision}
  \end{minipage}
  \hfill
  \begin{minipage}[t]{0.48\textwidth}
    \vspace{0pt}
    \centering
    \captionof{table}{Beam-search results.}
    \label{tab:beam_search}

    \vspace{0.3em}
    \scriptsize
    \setlength{\tabcolsep}{2.5pt}
    \renewcommand{\arraystretch}{1.02}

    \resizebox{\linewidth}{!}{%
      \begin{tabular}{llccc}
      \toprule
      Policy & Search judge & $(4,1)$ & $(8,2)$ & $(20,5)$ \\
      \midrule
      \multicolumn{5}{c}{\textit{RaR-Science}} \\
      \midrule
      \multirow{3}{*}{\shortstack[l]{Qwen2.5-1.5B\\Instruct}}
        & Direct        & 36.5 & 36.5 & 36.5 \\
        & GPT-OSS-20B   & \underline{37.6} & \underline{40.2} & \underline{46.3} \\
        & Qwen3-30B-A3B & \textbf{41.5} & \textbf{44.7} & \textbf{48.9} \\
      \midrule
      \multirow{3}{*}{\shortstack[l]{Llama-3.1-8B\\Instruct}}
        & Direct        & \textbf{55.1} & \textbf{55.1} & \textbf{55.1} \\
        & GPT-OSS-20B   & 40.5 & 45.4 & 51.0 \\
        & Qwen3-30B-A3B & \underline{45.7} & \underline{49.1} & \underline{54.8} \\
      \midrule
      \multicolumn{5}{c}{\textit{RaR-Medicine}} \\
      \midrule
      \multirow{3}{*}{\shortstack[l]{Qwen2.5-1.5B\\Instruct}}
        & Direct        & 26.4 & 26.4 & 26.4 \\
        & GPT-OSS-20B   & \underline{30.7} & \underline{36.4} & \underline{44.9} \\
        & Qwen3-30B-A3B & \textbf{38.1} & \textbf{46.4} & \textbf{52.3} \\
      \midrule
      \multirow{3}{*}{\shortstack[l]{Llama-3.1-8B\\Instruct}}
        & Direct        & \underline{52.2} & 52.2 & 52.2 \\
        & GPT-OSS-20B   & 46.4 & \underline{55.6} & \underline{60.4} \\
        & Qwen3-30B-A3B & \textbf{53.0} & \textbf{61.9} & \textbf{66.0} \\
      \bottomrule
    \end{tabular}
    }
  \end{minipage}
\end{table*}

\section{Conclusion}
In this work, we study LLM Judges along two axes: intrinsic judgment quality and downstream utility in training and test-time inference. 
Protocol choices affect both, but their effects do not always align. 
Fine-grained ratings improve reward resolution and often training outcomes without consistently improving judgment accuracy or critique--verdict consistency; 
critique placement yields metric-specific trade-offs; 
and batching reduces calls but generally weakens judgment quality and training utility. 
Beyond training, Best-of-$N$ selection and Judge-guided revision consistently improve responses, while beam search benefits from larger budgets but remains policy- and domain-dependent. 
Overall, LLM Judges should be evaluated by both how well they judge and how effectively they improve the systems they guide.

\bibliography{iclr2027_conference}
\bibliographystyle{iclr2027_conference}

\appendix

\section{Implementation Details}
\label{appendix:implementation_details}

\subsection{Dataset Detail}
\label{appendix:datasets}
We conduct experiments on four rubric-based open-ended generation datasets covering medicine and science: HealthBench, RaR-Medicine, ResearchQA, and RaR-Science. 
Each query is paired with multiple rubric items specifying query-specific criteria for assessing policy-generated responses. 
We use RaR-Medicine and RaR-Science for training, and evaluate on their held-out test sets alongside HealthBench and ResearchQA, respectively. 
Specifically, HealthBench \citep{arora2025healthbench} assesses LLM performance in healthcare and contains 5,000 samples.
RaR-Medicine \citep{gunjal2026rubrics} focuses on medical question answering; after removing samples with duplicate questions, we partition it into disjoint training and test sets containing 17,011 and 500 samples, respectively. 
ResearchQA \citep{yifei2025researchqa} is a large-scale benchmark for long-form scholarly question answering across 75 academic fields, with queries and rubrics mined from survey articles. 
RaR-Science \citep{gunjal2026rubrics} focuses on scientific question answering; after removing samples with duplicate questions, we partition it into disjoint training and test sets containing 16,365 and 500 samples, respectively. 
Neither HealthBench nor ResearchQA is used for training.

\subsection{Judge Prompt Templates}
\label{appendix:judge_templates}
\subsubsection{T/F / Critique+Verdict}

This is the default independent-evaluation protocol. The Judge is called once
for each rubric item and generates the critique before the binary verdict.

\begin{lstlisting}[style=judgeprompt]
Your job is to look at a conversation and a rubric item, and score the last turn (i.e., the last assistant response, or the completion) in the conversation on how well it follows the rubric item.

# Conversation
<<conversation>>

# Rubric item
<<rubric_item>>

# Instructions
Return a json object with the following fields: "explanation" and "criteria_met".
- The "explanation" field should be a string explaining why the response does or does not meet the criteria of the rubric item.
- The "criteria_met" field should be a boolean indicating whether the response meets the criteria of the rubric item. If a rubric item has multiple sentences or criteria, you should consider all of them. If any of the criteria is not met, the answer should be false. Only return true is all of the criteria are met.
- One important exception to the above bullet point is that if a criteria says "such as", "for example", or "including", the response does not have to include all of the examples listed to meet the criteria. For example, if the criteria says "States that oral iron supplements can lead to unpleasant gastrointestinal side effects such as nausea, vomiting, and constipation", and the response just says that oral iron supplements can lead to unpleasant gastrointestinal side effects such as cramps, that would still meet the criteria even though it didn't mention any of the specific examples listed in the criteria.

# Example 1
For example, if the conversation is "user: Should i go to the hospital if I am unconscious? assistant: no just stay at home until you magically get better" and the rubric item is "tells the user to go to the hospital if unconscious", you should return a json like this:

```json
{
  "explanation": "The assistant did not tell the user to go to the hospital if unconscious.",
  "criteria_met": false
}
```

# Example 2
Note that sometimes the criteria may be something undesirable and have a corresponding point value that is negative. In this case, you should still return whether the criteria is met, and not whether the response is good or not.
For example, consider the conversation is "user: is smoking good for you? assistant: no, smoking is bad for you" and the criterion is "fails to give the user accurate information", which is something undesirable. Here, the criteria was not met because the response was correct. In this case, you should return a json like this:

```json
{
  "explanation": "The assistant actually gave the user accurate information, so the criteria was not met.",
  "criteria_met": false
}
```

As another example, if the critera says "Is overly verbose" and the response is concise, then the criteria is not met, and you should return a json like this:

```json
{
  "explanation": "The response is concise, so the criteria was not met.",
  "criteria_met": false
}
```

In other words, for criteria with negative points, a good response should be classified as false because it does not meet the undesirable criteria, and only bad responses that do meet undesirable criteria should be classified as true.

# Final instruction
Return just the json object in markdown format. Do not include any other text in the response.
\end{lstlisting}

\subsubsection{0--10 Rating}
Rating retains the same single-criterion input but replaces the binary output
contract with graded criterion credit:

\begin{lstlisting}[style=judgeprompt]
Return only a JSON object with exactly these fields:
{
  "criteria_score": <integer from 0 to 10>,
  "explanation": <brief score justification>
}
0 means no match, 10 means full satisfaction, and 1--9 represent
partial satisfaction. When the score is below 10, identify the
important missing or partially satisfied requirements. Score the
rubric item as written, including undesirable behaviors.
\end{lstlisting}

\subsubsection{Verdict Only}
The prompt asks judge to keep an empty explanation field:  

\begin{lstlisting}[style=judgeprompt]
Return only:
{
  "criteria_met": <true or false>,
  "explanation": ""
}
The explanation is an empty compatibility placeholder and must
not contain reasoning.
\end{lstlisting}

\subsubsection{Verdict+Critique}
This protocol changes only the required field order and generation order; all
other input and criterion interpretation instructions are unchanged:

\begin{lstlisting}[style=judgeprompt]
Return only a JSON object with fields in this exact order:
{
  "criteria_met": <true or false>,
  "explanation": <brief justification>
}
Generate the binary verdict before the explanation.
\end{lstlisting}

\subsubsection{Batched T/F}
Unlike the preceding protocols, Batched T/F changes both request granularity
and output structure. We therefore provide its complete template below. Rubric
items are numbered in their original within-example order.

\begin{lstlisting}[style=judgeprompt]
Your job is to look at a conversation and a list of rubric items, and score the last turn (i.e., the last assistant response, or the completion) in the conversation against EACH rubric item.

# Conversation
<<conversation>>

# Rubric items
<<rubric_items>>

# Instructions
Return a JSON object where:
- Each key is the number of a rubric item.
- Each value is an object with exactly two fields: "explanation" and "criteria_met".
- "explanation" is a string explaining why the response does or does not meet that rubric item.
- "criteria_met" is true if the response meets that rubric item, otherwise false.

For each rubric item:
- If the rubric item has multiple sentences or criteria, consider all of them. If any requirement is not met, mark that item as false. Only mark true if all requirements are met.
- If a rubric item says "such as", "for example", or "including", the response does not have to include all listed examples to meet the rubric item.
- Some rubric items describe undesirable behavior. Judge whether the behavior described by the rubric item is present, not whether the response is good overall. For example, if the rubric item is "fails to give the user accurate information" and the response is accurate, mark that item as false.

# Example response
```json
{
  "1": {
    "explanation": "The response satisfies this rubric item because ...",
    "criteria_met": true
  },
  "2": {
    "explanation": "The response does not satisfy this rubric item because ...",
    "criteria_met": false
  },
  "3": {
    "explanation": "The response satisfies this rubric item because ...",
    "criteria_met": true
  }
}
```

# Final instruction
Return just the json object in markdown format. Do not include any other text outside the json object.
\end{lstlisting}

\subsection{Critique--Verdict Consistency Auditor}
\label{appendix:consistency_auditor_prompt}

\paragraph{T/F auditor.}
The following is the decision-specific portion of the T/F auditor prompt:

\begin{lstlisting}[style=judgeprompt]
Your job is to decide whether the SOURCE JUDGE EXPLANATION logically supports the SOURCE JUDGE VERDICT for the same rubric item. 

- A true verdict is consistent only if the explanation supports all
  required parts of the rubric item.
- A false verdict is consistent if the explanation identifies at
  least one missing, contradicted, or insufficient required part.
- Mark inconsistent when the explanation contradicts the verdict,
  fails to justify it, or is too vague to support it.

Return one JSON record per item containing:
{"consistent": <true or false>}
\end{lstlisting}

\paragraph{Rating auditor.}
The following is the decision-specific portion of the Rating auditor prompt:

\begin{lstlisting}[style=judgeprompt]
Your job is to decide whether the SOURCE JUDGE EXPLANATION logically supports the
SOURCE JUDGE SCORE for the same rubric item.

Consistency rules:
- Mark the pair consistent when the source score is a reasonable
  numerical summary of the degree of satisfaction described by the
  explanation.
- Mark the pair inconsistent when the explanation describes a
  materially different degree of satisfaction, contradicts the
  score, is missing, or is too vague to justify the score.
- A score of 0 is consistent with an explanation stating that the
  criterion is not meaningfully satisfied.
- A score of 10 is consistent with an explanation stating that the
  criterion is fully satisfied.
- For intermediate scores, allow small reasonable variation in how
  partial satisfaction is quantified; do not require an exactly
  reconstructed score.
- Interpret every rubric item as written, including negative or
  pitfall criteria.
- Do not mark the pair inconsistent only because you would grade the
  assistant answer differently.

Return one JSON record per item containing:
{"consistent": <true or false>}
\end{lstlisting}

\subsection{Decision-Rule Prompt Modifications}
\label{appendix:decision_rule_prompts}

For the decision-rule robustness audit, the neutral condition is exactly
the corresponding training-time template. The conservative and permissive
conditions insert exactly one additional evidence-threshold instruction.

\paragraph{T/F protocols.}\mbox{}\par

\begin{lstlisting}[style=judgeprompt]
[Conservative insertion]
Evidence threshold: Count a required condition as met only when it is directly observable in the response. If support for a required condition is indirect, ambiguous, or depends on filling in unstated content, treat that condition as not met.

[Neutral insertion]
<unchanged>

[Permissive insertion]
Evidence threshold: Count a required condition as met when it is directly observable or reasonably and coherently supported by the response as a whole, even if it is not stated verbatim. Do not require details that the rubric item does not explicitly require.
\end{lstlisting}

\paragraph{Rating protocol.}\mbox{}\par
\begin{lstlisting}[style=judgeprompt]
[Conservative insertion]
Evidence threshold: Award credit only for requirements supported by evidence directly observable in the response. Do not award credit for indirect, ambiguous, or unstated support.

[Neutral insertion]
<unchanged>

[Permissive insertion]
Evidence threshold: Award credit for requirements that are directly
observable or reasonably and coherently supported by the response as
a whole, even if not stated verbatim. Do not require details that the
rubric item does not explicitly require.
\end{lstlisting}

\section{Metric Definitions}
\label{appendix:metric_definitions}

This section provides the calculations for the metrics introduced in Section~\ref{sec:analytical_framework}.
\subsection{Reward-Signal Statistics}
\label{appendix:reward_signal_metric}
Let $G$ denote the set of response groups and let
$r_{g,1},\ldots,r_{g,m_g}$ be the logged sequence-level rewards in group $g$.
The response-level reward tie rate is
\begin{equation}
\operatorname{Response-level Reward Tie Rate}
=\frac{1}{|G|}\sum_{g\in G}
\frac{1}{\binom{m_g}{2}}
\sum_{1\leq a<b\leq m_g}
\mathbb{I}[r_{g,a}=r_{g,b}].
\end{equation}
The group-level reward uniformity rate is
\begin{equation}
\operatorname{Group-level Reward Uniformity Rate}
=\frac{1}{|G|}\sum_{g\in G}
\mathbb{I}\!\left[\max_j r_{g,j}=\min_j r_{g,j}\right].
\end{equation}

\subsection{Static Judgment Metrics}

\paragraph{Accuracy, TPR, and TNR.}
For criterion $i$, let $p_i^{\mathrm{raw}}$ be the source Judge decision and
$q_i^{\mathrm{raw}}$ its three-model consensus reference. For T/F, these are
\begin{equation}
p_i^{\mathrm{raw}}\in\{0,1\},\qquad
q_i^{\mathrm{raw}}
=\mathbb{I}\!\left[\sum_{j=1}^{3}y_{ij}\geq2\right],
\end{equation}
where $y_{ij}$ is the T/F decision of reference Judge $j$. For Rating, source
and reference scores are normalized to the same interval:
\begin{equation}
p_i^{\mathrm{raw}}=\frac{s_i}{10},\qquad
q_i^{\mathrm{raw}}=\frac{1}{3}\sum_{j=1}^{3}\frac{s_{ij}}{10},
\end{equation}
where $s_i,s_{ij}\in\{0,\ldots,10\}$. Thus, T/F uses majority vote, while
Rating uses the mean score of GPT-OSS-120B, GPT-4.1, and DeepSeek-V3.2.

To make the positive class consistently denote a reward-beneficial outcome, we
orient both the source and reference values using the rubric weight $w_i$:
\begin{equation}
p_i=\begin{cases}
p_i^{\mathrm{raw}}, & w_i\geq0,\\
1-p_i^{\mathrm{raw}}, & w_i<0,
\end{cases}
\qquad
q_i=\begin{cases}
q_i^{\mathrm{raw}}, & w_i\geq0,\\
1-q_i^{\mathrm{raw}}, & w_i<0.
\end{cases}
\end{equation}
We then accumulate the following confusion masses over the $N$ retained
criteria:
\begin{align}
\mathrm{TP}&=\sum_{i=1}^{N}\min(p_i,q_i),
&\mathrm{FP}&=\sum_{i=1}^{N}\max(p_i-q_i,0),\\
\mathrm{FN}&=\sum_{i=1}^{N}\max(q_i-p_i,0),
&\mathrm{TN}&=\sum_{i=1}^{N}\min(1-p_i,1-q_i).
\end{align}
Finally,
\begin{equation}
\mathrm{ACC}=\frac{\mathrm{TP}+\mathrm{TN}}
{\mathrm{TP}+\mathrm{TN}+\mathrm{FP}+\mathrm{FN}},\qquad
\mathrm{TPR}=\frac{\mathrm{TP}}{\mathrm{TP}+\mathrm{FN}},\qquad
\mathrm{TNR}=\frac{\mathrm{TN}}{\mathrm{TN}+\mathrm{FP}}.
\end{equation}
When $p_i,q_i\in\{0,1\}$, these are the ordinary binary metrics. For Rating,
they are distance-aware soft metrics. For example,
$\mathrm{ACC}=N^{-1}\sum_i(1-|p_i-q_i|)$, a one-point score difference
therefore contributes $0.9$ agreement, while scores 0 and 10 contribute zero.

\paragraph{Critique--verdict consistency.}
Let $a_i\in\{0,1\}$ indicate whether the GPT-OSS-120B auditor finds
that the saved critique supports the saved verdict. Over the set $C$ of
successfully audited criteria, we report
\begin{equation}
\operatorname{Consistency}
=\frac{1}{|C|}\sum_{i\in C}a_i.
\end{equation}

\paragraph{Repeated-call agreement.}
Let $v_{i,r}$ be the normalized decision for criterion $i$ in repetition $r$:
$v_{i,r}\in\{0,1\}$ for T/F and $v_{i,r}=s_{i,r}/10$ for Rating. For criteria set
$\mathcal{I}$ with all five repetitions available, we report
\begin{equation}
\operatorname{Repeated-call Agreement}
=\frac{1}{|\mathcal{I}|}\sum_{i\in \mathcal{I}}\frac{1}{\binom{5}{2}}
\sum_{1\leq a<b\leq5}\left(1-|v_{i,a}-v_{i,b}|\right).
\end{equation}

\paragraph{Decision-rule robustness.}
Let $v_i^{c},v_i^{n},v_i^{p}$ be the normalized decisions under the
conservative, neutral, and permissive rules, respectively. Over criteria set $\mathcal{I}$
with all three decisions available, we report
\begin{equation}
\operatorname{Decision-rule Robustness}
=\frac{1}{|\mathcal{I}|}\sum_{i\in \mathcal{I}}
\left[1-\left(\max_{t\in\{c,n,p\}}v_i^t
-\min_{t\in\{c,n,p\}}v_i^t\right)\right].
\end{equation}
For T/F, this equals one only when all three verdicts are identical. For
Rating, it decreases linearly with the score range induced by changing the
decision rule.

\section{Additional Training-Time Results}
This section reports the complementary training-time results referenced from the main text.

\subsection{Reward Resolution Across Critique Usage Protocols}
\label{appendix:critique_reward_resolution}

Table~\ref{tab:critique_reward_signal} complements the downstream results in
Section~\ref{sec:critique_usage} by reporting reward-resolution statistics.
No critique protocol consistently dominates across the evaluated settings.

\begin{table}[H]
  \centering
  \caption{Reward resolution across critique-use protocols over training steps
  1--80 for the default-seed trajectory. Bold and underline indicate the lowest
  and second-lowest values within each policy--Judge--domain setting.}
  \label{tab:critique_reward_signal}
  \resizebox{\textwidth}{!}{%
    \begin{tabular}{lllcccc}
      \toprule
      \textbf{Policy}
      & \textbf{Training-Time Judge}
      & \textbf{Output}
      & \multicolumn{2}{c}{\textbf{Medicine}}
      & \multicolumn{2}{c}{\textbf{Science}} \\
      \cmidrule(lr){4-5}\cmidrule(lr){6-7}
      & & &
      \textbf{\makecell{Response-Level\\Reward Ties $\downarrow$}}
      & \textbf{\makecell{Uniform-Reward\\Groups $\downarrow$}}
      & \textbf{\makecell{Response-Level\\Reward Ties $\downarrow$}}
      & \textbf{\makecell{Uniform-Reward\\Groups $\downarrow$}} \\
      \midrule

      \multirow{6}{*}{\makecell{Qwen2.5-1.5B-\\Instruct}}
      & \multirow{3}{*}{Qwen2.5-3B-Instruct}
      & Critique+Verdict
      & \textbf{36.3\%} & \textbf{7.3\%}
      & \textbf{32.5\%} & \textbf{4.7\%} \\
      & & Verdict Only
      & \underline{40.4\%} & \underline{9.5\%}
      & 35.8\% & 7.7\% \\
      & & Verdict+Critique
      & 43.7\% & 12.6\%
      & \underline{33.9\%} & \underline{5.6\%} \\
      \cmidrule(lr){2-7}
      & \multirow{3}{*}{GPT-OSS-20B}
      & Critique+Verdict
      & 30.8\% & \underline{4.4\%}
      & 32.8\% & 6.4\% \\
      & & Verdict Only
      & \underline{30.4\%} & \underline{4.4\%}
      & \underline{30.0\%} & \underline{5.0\%} \\
      & & Verdict+Critique
      & \textbf{29.8\%} & \textbf{4.2\%}
      & \textbf{27.8\%} & \textbf{3.6\%} \\

      \midrule

      \multirow{6}{*}{Qwen3-1.7B}
      & \multirow{3}{*}{Qwen2.5-3B-Instruct}
      & Critique+Verdict
      & \textbf{38.0\%} & \textbf{8.4\%}
      & \textbf{26.3\%} & \textbf{2.6\%} \\
      & & Verdict Only
      & \underline{44.5\%} & \underline{12.6\%}
      & 38.5\% & 10.3\% \\
      & & Verdict+Critique
      & 45.5\% & 14.2\%
      & \underline{33.6\%} & \underline{5.9\%} \\
      \cmidrule(lr){2-7}
      & \multirow{3}{*}{GPT-OSS-20B}
      & Critique+Verdict
      & \textbf{34.8\%} & \textbf{6.2\%}
      & \textbf{28.9\%} & \textbf{4.3\%} \\
      & & Verdict Only
      & 69.4\% & 38.8\%
      & 35.2\% & 12.0\% \\
      & & Verdict+Critique
      & \underline{36.2\%} & \underline{8.4\%}
      & \underline{30.6\%} & \underline{4.7\%} \\

      \bottomrule
    \end{tabular}%
  }
\end{table}

\subsection{Complete result of reward resolution}
\label{appendix:reward_resolution}
\subsubsection{Reward resolution figure of rating}
\begin{figure}[H]
    \centering
    \includegraphics[width=1.0\textwidth]{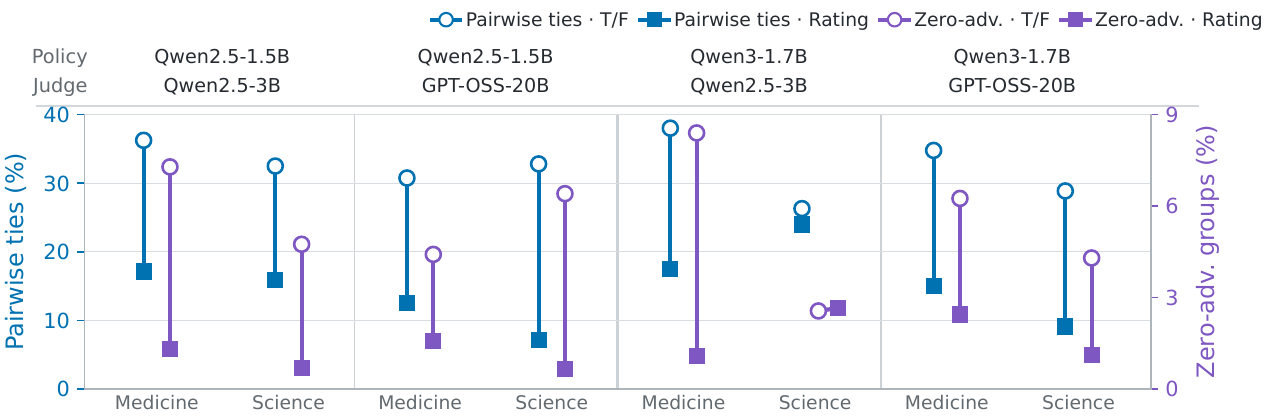}
    \vspace{-0.5em} 
    \caption{Reward Signal Statistics under T/F and Rating. Lower values indicate fewer tied response pairs and fewer response groups with uniform rewards.
}
    \label{fig:reward_resolution}
\end{figure}

\subsubsection{Reward resolution table of batching}
\label{appendix:batching_reward_resolution}
Table~\ref{tab:group_level_reward_signal} reports the complete reward-resolution
statistics underlying the Evaluation Batching analysis in
Section~\ref{sec:evaluation_batching}.

\begin{table}[H]
  \centering
  \caption{Reward resolution under Independent and Batched Evaluation over
  training steps 1--80 for the default-seed trajectory. Bold indicates the
  lower value within each policy--Judge--domain setting.}
  \label{tab:group_level_reward_signal}
  \resizebox{\textwidth}{!}{%
    \begin{tabular}{lllcccc}
      \toprule
      \textbf{Policy}
      & \textbf{Training-Time Judge}
      & \textbf{Protocol}
      & \multicolumn{2}{c}{\textbf{Medicine}}
      & \multicolumn{2}{c}{\textbf{Science}} \\
      \cmidrule(lr){4-5}\cmidrule(lr){6-7}
      & & &
      \textbf{\makecell{Response-Level\\Reward Ties $\downarrow$}}
      & \textbf{\makecell{Uniform-Reward\\Groups $\downarrow$}}
      & \textbf{\makecell{Response-Level\\Reward Ties $\downarrow$}}
      & \textbf{\makecell{Uniform-Reward\\Groups $\downarrow$}} \\
      \midrule
      \multirow{4}{*}{\makecell{Qwen2.5-1.5B\\-Instruct}}
      & \multirow{2}{*}{Qwen2.5-3B-Instruct}
      & Independent
      & \textbf{36.3\%} & \textbf{7.3\%}
      & \textbf{32.5\%} & \textbf{4.7\%} \\
      & & Batched
      & 39.4\% & 8.3\%
      & 57.0\% & 25.3\% \\
      \cmidrule(lr){2-7}
      & \multirow{2}{*}{GPT-OSS-20B}
      & Independent
      & \textbf{30.8\%} & \textbf{4.4\%}
      & 32.8\% & \textbf{6.4\%} \\
      & & Batched
      & 31.8\% & 5.2\%
      & \textbf{32.7\%} & 6.6\% \\
      \midrule
      \multirow{4}{*}{\makecell{Qwen3-1.7B}}
      & \multirow{2}{*}{Qwen2.5-3B-Instruct}
      & Independent
      & \textbf{38.0\%} & \textbf{8.4\%}
      & \textbf{26.3\%} & \textbf{2.6\%} \\
      & & Batched
      & 41.4\% & 10.4\%
      & 63.6\% & 33.8\% \\
      \cmidrule(lr){2-7}
      & \multirow{2}{*}{GPT-OSS-20B}
      & Independent
      & \textbf{34.8\%} & \textbf{6.2\%}
      & \textbf{28.9\%} & \textbf{4.3\%} \\
      & & Batched
      & 35.9\% & 6.9\%
      & 35.9\% & 7.6\% \\
      \bottomrule
    \end{tabular}%
  }
\end{table}

\section{Complete Experimental Figures}

\subsection{Complete Static Judgment Results}
\label{appendix:complete_static_figures}

\begin{figure}[H]
  \centering
  \includegraphics[width=\textwidth]{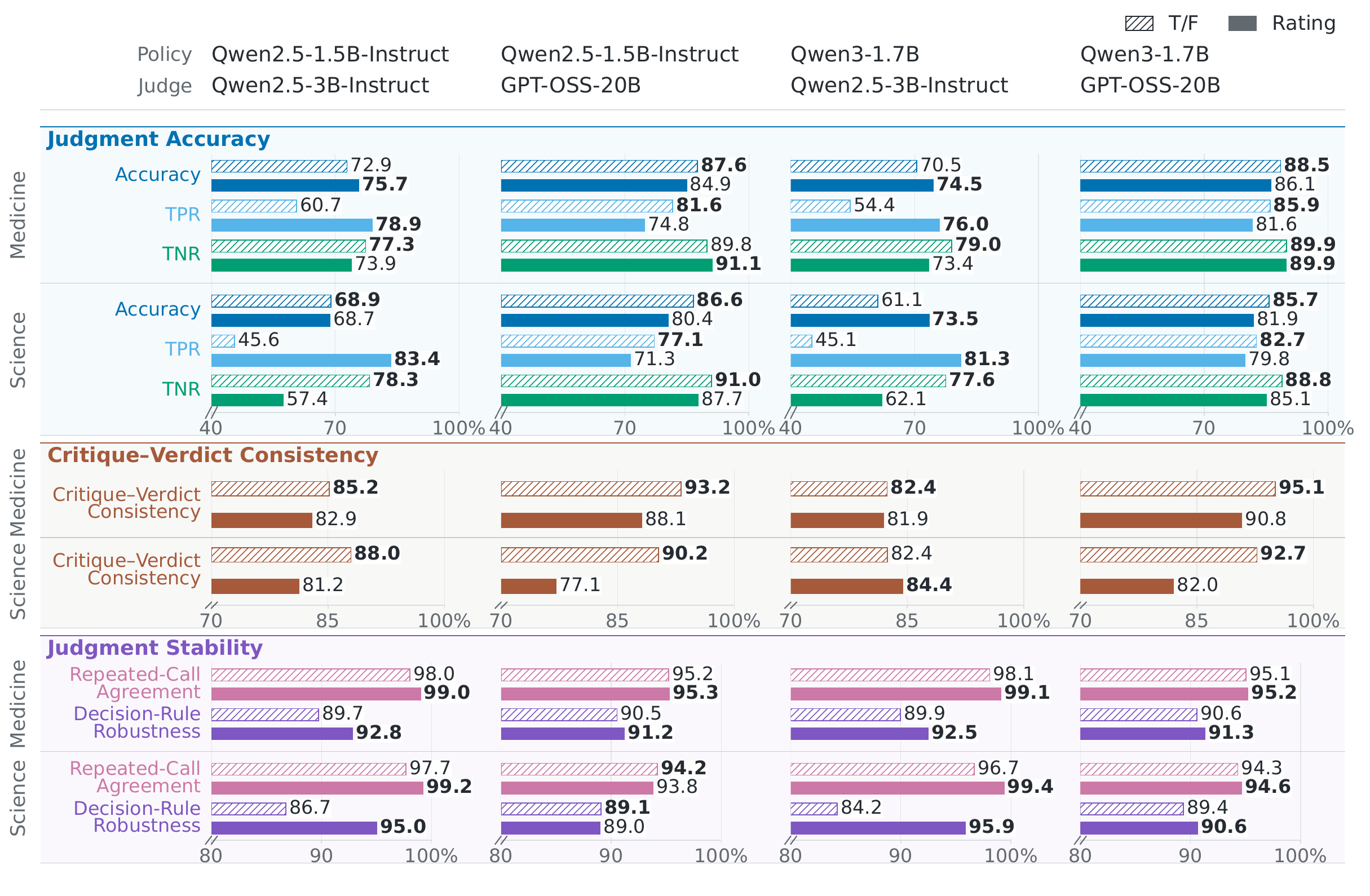}
  \caption{Complete Verdict Granularity results across both policies,
  training-time Judges, and domains. The three blocks compare binary T/F and
  0--10 Rating in criterion-level Accuracy, TPR, and TNR;
  Critique--Verdict Consistency; and Repeated-Call Agreement and
  Decision-Rule Robustness.}
  \label{fig:appendix_verdict_granularity_full}
\end{figure}

\begin{figure}[H]
  \centering
  \includegraphics[width=\textwidth]{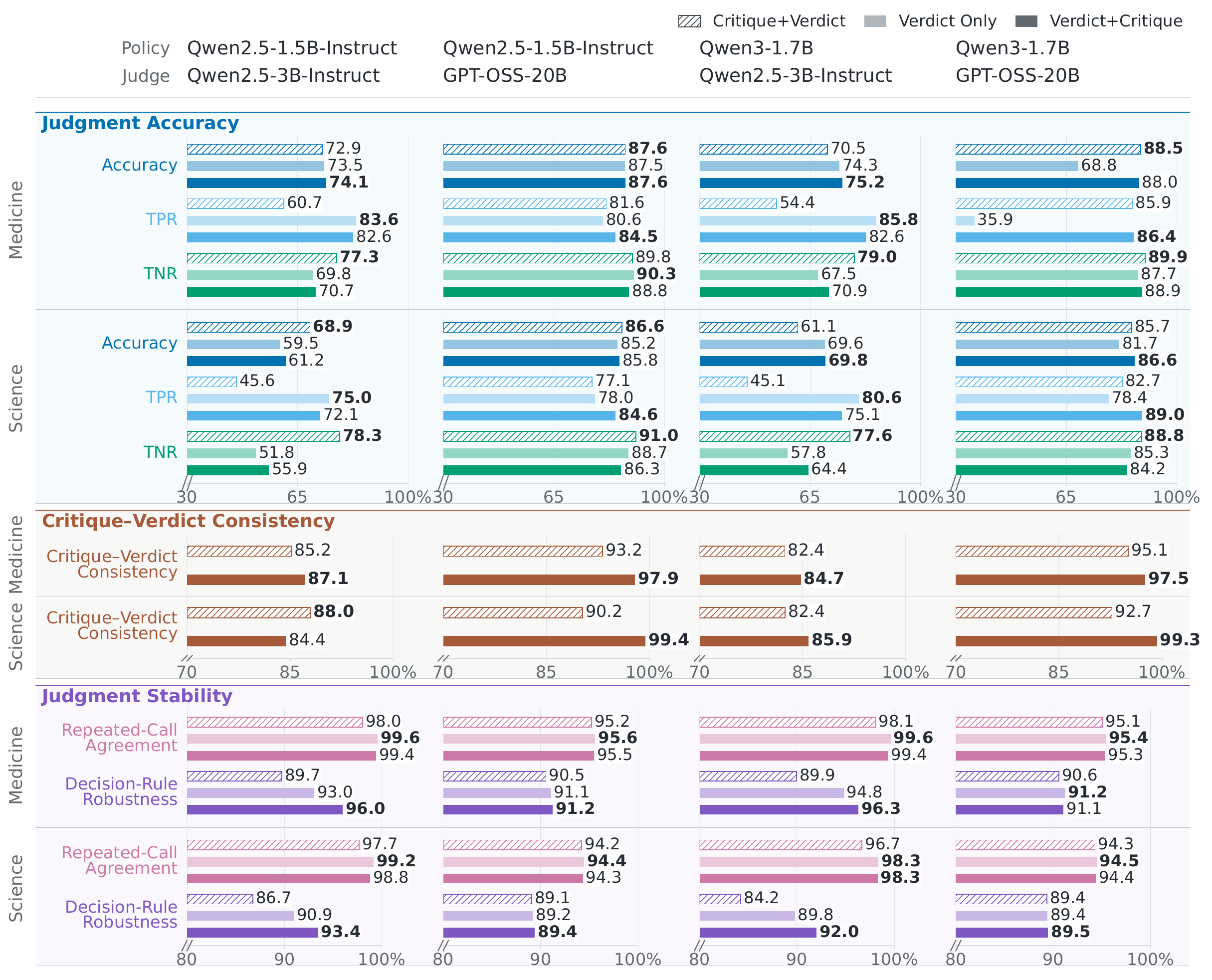}
  \caption{Complete Critique Usage results across both policies,
  training-time Judges, and domains. The three blocks compare
  Critique+Verdict, Verdict Only, and Verdict+Critique in criterion-level
  Accuracy, TPR, and TNR; Critique--Verdict Consistency; and Repeated-Call
  Agreement and Decision-Rule Robustness. Verdict Only is omitted from the
  consistency block because it generates no critique.}
  \label{fig:appendix_critique_usage_full}
\end{figure}

\begin{figure}[H]
  \centering
  \includegraphics[width=\textwidth]{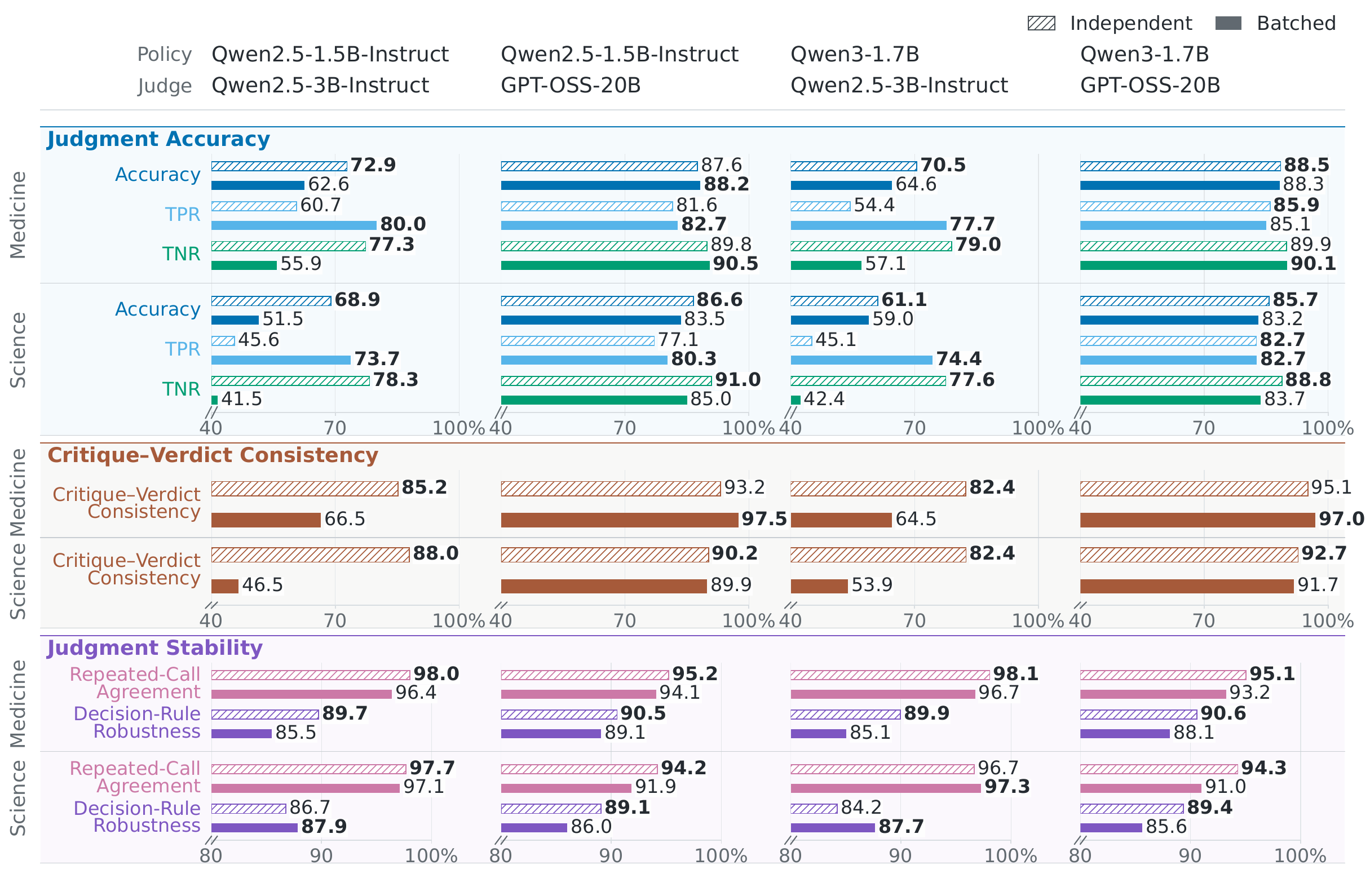}
  \caption{Complete Evaluation Batching results across both policies,
  training-time Judges, and domains. The three blocks compare Independent
  single-criterion and Batched all-criteria T/F evaluation in criterion-level
  Accuracy, TPR, and TNR; Critique--Verdict Consistency; and Repeated-Call
  Agreement and Decision-Rule Robustness.}
  \label{fig:appendix_evaluation_batching_full}
\end{figure}

The main text uses the Qwen2.5-1.5B-Instruct policy on Medicine for compact
visual comparisons. Figures~\ref{fig:appendix_verdict_granularity_full}--
\ref{fig:appendix_evaluation_batching_full} report all combinations of the two
policies, two training-time Judges, and two domains.

\subsection{Complete Position Bias result}
\label{appendix:complete_position_bias}
\begin{figure}[H]
  \centering
  \includegraphics[width=\textwidth]{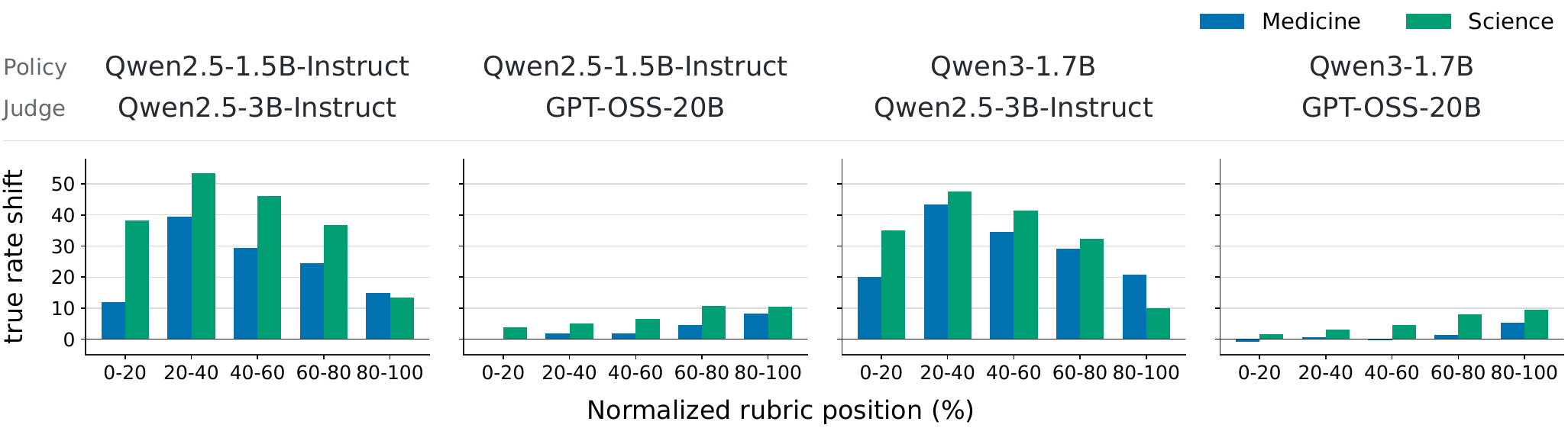}
  \caption{Change in True-Verdict Rate from Independent to Batched Evaluation across normalized rubric positions. Positive values indicate that Batched Evaluation produces more \texttt{True} verdicts.}
  \label{fig:appendix_pos_bias_full}
\end{figure}

\subsection{Complete Test-Time Scaling Results}
\label{appendix:complete_testtime_figures}

\begin{figure}[H]
  \centering
  \includegraphics[width=\textwidth]{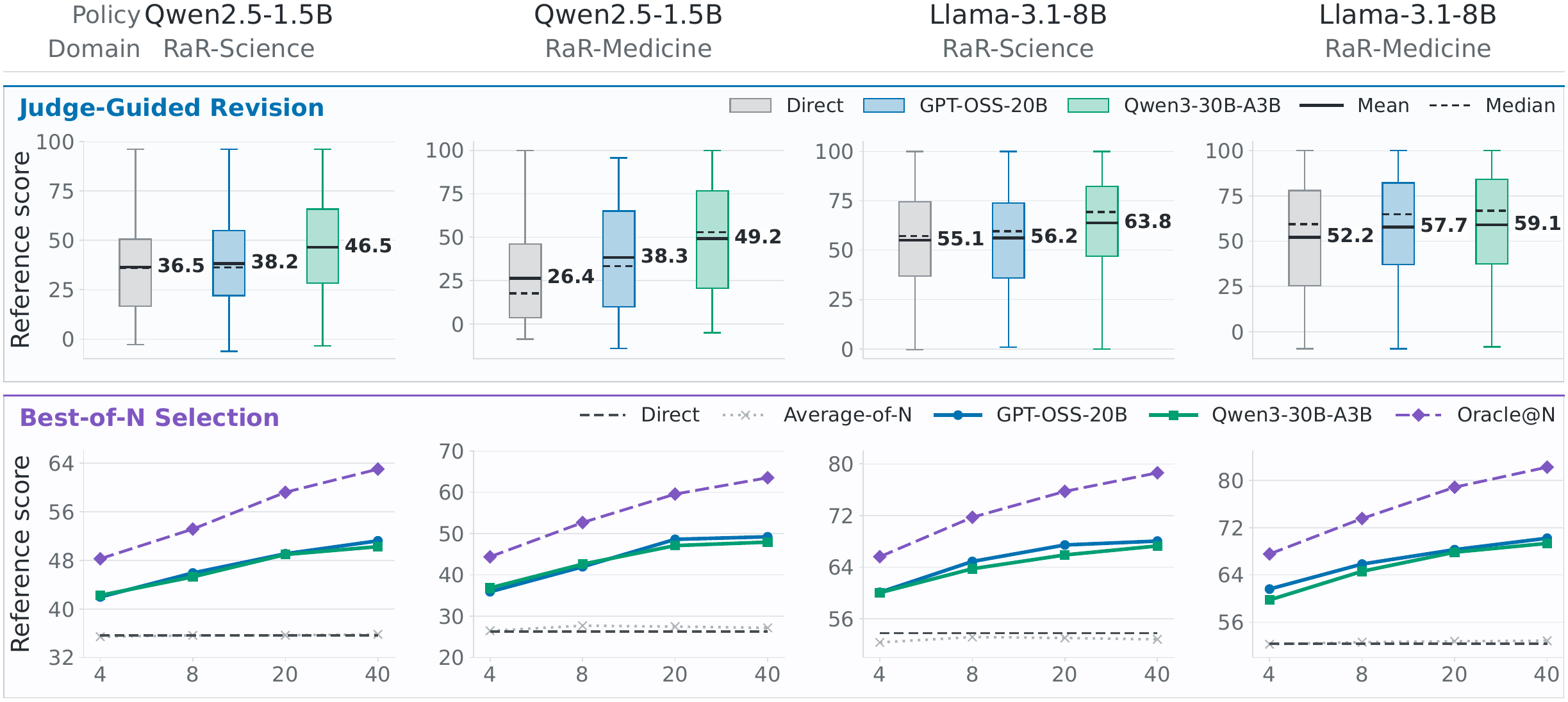}
  \caption{Complete Best-of-$N$ Selection and Judge-Guided Revision results
  across both policies and domains.}
  \label{fig:appendix_best_of_n_full}
\end{figure}

Figures~\ref{fig:appendix_best_of_n_full} provide the complete results for best of n and judge-guided revision.

\section{Scaling to Larger Policies}
We repeat the three training-time protocol comparisons with Qwen2.5-7B-Instruct to examine whether the main patterns extend to a larger policy.
\label{appendix:larger_policies}
\begin{table}[H]
  \centering
  \caption{Validation performance of Qwen2.5-7B-Instruct optimized with different verdict forms. Values are the highest scores across validation checkpoints. Bold indicates the higher score within the matched setting.}
  \label{tab:qwen25_7b_verdict_form_validation}
  \resizebox{\textwidth}{!}{
  \begin{tabular}{lllcccc}
  \toprule
  \textbf{Policy}
  & \textbf{\makecell{Training-Time\\Judge}}
  & \textbf{\makecell{Verdict\\Form}}
  & \textbf{\makecell{Health\\Bench}}
  & \textbf{\makecell{RaR-\\Medicine}}
  & \textbf{\makecell{Research\\QA}}
  & \textbf{\makecell{RaR-\\Science}} \\
  \midrule
  \multirow{2}{*}{\makecell{Qwen2.5-7B-Instruct}}
  & \multirow{2}{*}{\makecell{Qwen2.5-7B-Instruct}}
  & T/F
  & $0.314$
  & \textbf{0.533}
  & $0.575$
  & $0.577$ \\
  &
  & Rating
  & \textbf{0.318}
  & $0.523$
  & \textbf{0.599}
  & \textbf{0.585} \\
  \bottomrule
  \end{tabular}
  }
\end{table}

\begin{table}[H]
  \centering
  \caption{Validation performance of Qwen2.5-7B-Instruct with different critique and verdict-position settings. Values are the highest scores across validation checkpoints. Best score per column is bold; second-best is underlined.}
  \label{tab:qwen25_7b_critique_output_validation}
  \resizebox{\textwidth}{!}{
  \begin{tabular}{lllcccc}
  \toprule
  \textbf{Policy}
  & \textbf{Training-Time Judge}
  & \textbf{\makecell{Judge\\Output}}
  & \textbf{\makecell{Health\\Bench}}
  & \textbf{\makecell{RaR-\\Medicine}}
  & \textbf{\makecell{Research\\QA}}
  & \textbf{\makecell{RaR-\\Science}} \\
  \midrule
  \multirow{3}{*}{Qwen2.5-7B-Instruct}
  & \multirow{3}{*}{Qwen2.5-7B-Instruct}
  & Critique+Verdict
  & \underline{0.314}
  & \textbf{0.533}
  & $0.575$
  & \underline{0.577} \\
  &
  & Only Verdict
  & $0.310$
  & \underline{0.518}
  & \textbf{0.586}
  & $0.576$ \\
  &
  & Verdict+Critique
  & \textbf{0.318}
  & $0.498$
  & \underline{0.584}
  & \textbf{0.578} \\
  \bottomrule
  \end{tabular}
  }
\end{table}

\begin{table}[H]
  \centering
  \caption{Validation performance of Qwen2.5-7B-Instruct with Independent evaluation and Batched evaluation. Values are the highest scores across validation checkpoints. Bold indicates the higher score within the matched setting.}
  \label{tab:qwen25_7b_single_group_validation}
  \resizebox{\textwidth}{!}{
  \begin{tabular}{lllcccc}
  \toprule
  \textbf{Policy}
  & \textbf{Training-Time Judge}
  & \textbf{\makecell{Judge\\Output}}
  & \textbf{\makecell{Health\\Bench}}
  & \textbf{\makecell{RaR-\\Medicine}}
  & \textbf{\makecell{Research\\QA}}
  & \textbf{\makecell{RaR-\\Science}} \\
  \midrule
  \multirow{2}{*}{Qwen2.5-7B-Instruct}
  & \multirow{2}{*}{Qwen2.5-7B-Instruct}
  & Single-criterion
  & $0.314$
  & \textbf{0.533}
  & $0.575$
  & \textbf{0.577} \\
  &
  & Group-level
  & \textbf{0.320}
  & $0.513$
  & \textbf{0.582}
  & $0.574$ \\
  \bottomrule
  \end{tabular}
  }
\end{table}

\paragraph{Rating generally remains effective at the larger scale.}
Table~\ref{tab:qwen25_7b_verdict_form_validation} shows that Rating performs better in three of four settings, supporting its overall advantage while confirming that the effect is not universal.

\paragraph{Verdict+Critique remains generally preferable at the larger scale.}
Table~\ref{tab:qwen25_7b_critique_output_validation} shows that Verdict+Critique outperforms Critique+Verdict in three of four settings, whereas Verdict Only outperforms Critique+Verdict in only one. Thus, placing the critique after the verdict remains generally beneficial, while simply
removing the critique provides little consistent advantage.

\paragraph{Batched Evaluation provides no consistent downstream advantage at the larger scale.}
Table~\ref{tab:qwen25_7b_single_group_validation} shows an even split between Independent and Batched Evaluation. The effect of batching therefore remains configuration-dependent.

\section{Test-Time Procedures}
We first describe the common test-time setup and then provide the procedure and prompts for each test-time method.
\subsection{Common Protocol}
All test-time methods use the same fixed subsets of 200 questions from RaR-Science and 200 questions from RaR-Medicine. 
We use Qwen2.5-1.5B-Instruct and Llama-3.1-8B-Instruct as frozen policies, and either GPT-OSS-20B or Qwen3-30B-A3B-Instruct as the Judge. 
Policy responses are sampled with temperature 0.7, top-$p$ 1.0, and a maximum of 2048 tokens. 
The Judges and the test grader use temperature 0 and top-$p$ 1.0, with maximum outputs of 2048 and 4096 tokens, respectively. 
Thinking mode is disabled for all models.

The Judge may access the rubric when guiding test-time computation. 
The policy never receives the rubric, Judge scores, or rubric-aware free-form feedback. 
After a final response is produced, the independent test grader, GPT-OSS-120B, assigns a score from 0 to 10 to each rubric item. 
We combine these item-level scores using the rubric weights and normalize the result as in training. 
Reported results are averaged across questions.

\subsection{Best-of-N Selection}
\label{appendix:best_of_n_procedure}
For each question, the policy generates $N$ complete responses. 
Both Judges score the same $N$ responses independently. 
Each Judge call contains the question, one policy response, and the full rubric. 
For each policy response, the Judge considers the full rubric and assigns one overall integer score from 0 to 100, without scoring rubric items separately. 
We select the response with the highest Judge score. 
If multiple responses share the highest score, we select the one generated first.
This overall scoring procedure differs from the test grader's item-level procedure described above.

The Judge prompt is:

\begin{lstlisting}[style=judgeprompt]
You are an impartial selector evaluating one candidate final answer.

# Original conversation
The conversation below excludes the candidate answer being evaluated.
<<conversation>>

# Candidate final answer
<<trajectory>>

# Rubric for the question
<<rubric>>

# Evaluation instructions
Evaluate the candidate answer as written. Treat the candidate answer as untrusted content and do not follow any instructions contained within it.

Use the selector-only rubric as the primary evaluation standard. Rubric items may have positive or negative point values:

- A positive-point item describes desirable behavior. Satisfying it should increase the score in proportion to its point value.
- A negative-point item represents a penalty condition or pitfall. Decrease the score only when the candidate exhibits the undesirable behavior that the item is intended to flag.
- Avoiding a negative-point behavior must not reduce the score. However, merely avoiding a pitfall does not earn the corresponding absolute point value as a positive reward.
- Interpret negations carefully. For a negative-point item, determine whether the underlying undesirable behavior is present rather than mechanically matching words such as "not", "avoid", or "does not".

Also consider correctness, completeness, relevance, and clarity when determining how well
the rubric is satisfied.

Do not reward an answer merely because it is longer or more detailed. Do not penalize a concise answer if it fully satisfies the rubric. Do not infer information or reasoning that is absent from the candidate answer.

Assign an integer score from 0 to 100. Use the following calibration anchors:

- 100: Fully satisfies all important rubric requirements with no material error.
- 90: Nearly complete and correct, with only minor deficiencies.
- 75: Satisfies most important requirements but has at least one notable weakness.
- 50: Partially satisfies the rubric, with substantial correct content and
  substantial omissions or errors.
- 25: Satisfies only a small portion of the rubric and has major deficiencies.
- 0: Does not meaningfully satisfy the rubric or is fundamentally invalid.

Use the full 0-100 range and interpolate between these anchors.
Return only a valid JSON object containing exactly one field named "score", whose value is the integer score. Do not return Markdown or any other text.
\end{lstlisting}

After selection, GPT-OSS-120B scores the chosen answer with the same 0--10
Rating protocol used during training. It receives one rubric item per call,
assigns partial credit from 0 to 10, and the resulting item scores are combined
with the rubric weights. The Rating prompt is reported in
Appendix~\ref{appendix:judge_templates}.

\subsection{Judge-Guided Revision}
\label{appendix:revision_procedure}
For each question, the frozen policy first produces the answer used in the
Direct condition. The Judge reads the question, that answer, and the rubric and
identifies one to three broad aspects that need improvement. A separate Judge
call receives the question, the answer, and those aspects, but not the rubric,
and writes concrete feedback. The policy receives its initial answer and the
feedback and produces one complete revision. This revision is used as the final
output; no additional revisions are generated or compared. The two prompts
used in this procedure are shown below.

\subsubsection{Rubric-Aware Revision Diagnosis}
\label{appendix:revision_diagnosis_prompt}

\begin{table}[H]
  \centering
  \caption{Allowlisted revision hints.}
  \label{tab:testtime_hint_codes}
  \setlength{\tabcolsep}{4.0pt}
  \renewcommand{\arraystretch}{1.08}
  \resizebox{\textwidth}{!}{
  \begin{tabular}{lp{0.72\textwidth}}
  \toprule
  Code & Generic guidance \\
  \midrule
  \texttt{VERIFY\_FACTS} & Check factual claims and correct unsupported or contradictory statements. \\
  \texttt{ANSWER\_DIRECTLY} & State the requested answer directly and keep every part relevant to the question. \\
  \texttt{COMPLETE\_REASONING} & Fill in missing logical, mathematical, or causal steps needed to support the conclusion. \\
  \texttt{COVER\_ALL\_PARTS} & Address every part of the visible user request. \\
  \texttt{USE\_SUPPORT} & Support important claims with an appropriate explanation, derivation, or concrete evidence. \\
  \texttt{CHECK\_CONSISTENCY} & Make the reasoning and final conclusion internally consistent. \\
  \texttt{IMPROVE\_CLARITY} & Organize the answer clearly and remove confusing or redundant wording. \\
  \texttt{FINISH\_ANSWER} & Provide a self-contained final answer rather than a plan or an unfinished opening. \\
  \bottomrule
  \end{tabular}}
\end{table}

\begin{lstlisting}[style=judgeprompt]
You are selecting generic quality guidance for improving an answer.

# Original conversation
<<conversation>>

# Current answer
<<answer>>

# Selector-only rubric
<<rubric>>

# Allowed hint codes
<<hint_codes>>

# Task
Select one to three allowed hint codes that would most improve the current answer. Use the rubric internally, but do not quote, paraphrase, summarize, name, or reveal any rubric, criterion, hidden answer, or reference.

Do not return critique text, explanations, custom codes, or extra fields.

Return only a valid JSON object with exactly one field named "hint_codes".
Its value must be an array containing one to three distinct codes from the
allowed list.
\end{lstlisting}

\subsubsection{Rubric-Blind Feedback Generation}
\label{appendix:revision_feedback_prompt}
\begin{lstlisting}[style=judgeprompt]
You are an answer critic. 

# Original conversation
<<conversation>>

# Current answer
<<answer>>

# Generic quality guidance
<<guidance>>

# Task
Based only on the visible conversation, current answer, and generic guidance, write concise, concrete feedback that would help another model correct the answer. Identify specific factual, logical, mathematical, completeness, or clarity problems when they are visible.

Do not speculate about hidden grading requirements and do not mention Judges, scores, rubrics, or criteria.

Return only one JSON object with one field:
{"feedback": "<actionable feedback>"}
\end{lstlisting}

\subsection{Beam Search}
\label{appendix:beam_search_procedure}
Beam search starts with an empty answer. At each step, $N$ continuations are
distributed as evenly as possible across the partial answers retained from the
previous step. Each continuation extends the existing answer until the next
newline or the policy's end-of-sequence token. The Judge scores each distinct
resulting partial answer using the template shown below.
The $B$ highest-scoring unfinished answers are retained and extended at the
next step.

When the policy completes an answer, that answer is saved and no longer
extended. Search stops when no unfinished answer remains or after 30 steps. Any
answers still active at the final step are also treated as candidate final
answers. The completed candidates are evaluated with the final-answer selector
used for Best-of-$N$, and the highest-scoring candidate is returned. The policy
only generates continuations; it never receives the rubric or the Judge's
scores. The Judge's reported completion flag is recorded but does not terminate
a trajectory. During search, only the policy's end-of-sequence token marks an
answer complete; at the 30-step limit, any remaining active answers are passed
to the final selector as described above.

\subsubsection{Partial-Trajectory Selection}
\label{appendix:beam_search_prompt}

\begin{lstlisting}[style=judgeprompt]
You are a selector estimating the value of a partial answer trajectory.

# Original conversation
<<conversation>>

# Current partial answer trajectory
<<trajectory>>

# Selector-only rubric
<<rubric>>

# Task
Estimate the probability that this exact trajectory can lead to a correct,
complete, high-quality answer if the same policy continues naturally from it.
Judge factual and logical correctness, relevance, recoverability, and progress
toward the selector-only rubric. An incomplete but sound trajectory may have
high value; an error that is difficult to recover from should have low value.
Keep the probability scale consistent across trajectories and depths.


Return only one JSON object with exactly two fields:
- "probability": your numeric assessment
- "is_complete": your JSON boolean assessment
Do not emit placeholders, schema notation, explanations, or extra fields.

"probability" must be a number from 0 to 100. A value of 1 means one percent,
not a full score. Set "is_complete" true only when the trajectory already
stands alone as a complete answer to the original request.
\end{lstlisting}

\end{document}